\documentclass{article}

\PassOptionsToPackage{numbers, compress}{natbib}

\usepackage[eandd, final]{neurips_2026}

\usepackage[utf8]{inputenc} 
\usepackage[T1]{fontenc}    
\usepackage{hyperref}       
\usepackage{color}
\usepackage{url}            
\usepackage{booktabs}       
\usepackage{amsfonts}       
\usepackage{amsmath}        
\usepackage{nicefrac}       
\usepackage{microtype}      
\usepackage[table]{xcolor}  
\usepackage{graphicx}       
\usepackage{siunitx}
\usepackage{pifont}
\newcommand{\cmark}{\textcolor{green!80!black}{\ding{51}}} 
\newcommand{\xmark}{\textcolor{red}{\ding{55}}}

\usepackage{wasysym}

\usepackage{multirow}
\usepackage{makecell}
\usepackage{colortbl}
\usepackage{array}
\usepackage{adjustbox}

\usepackage{tabularx}
\usepackage{subcaption}

\hypersetup{
    urlcolor=blue
}

\definecolor{groupyellow}{HTML}{FBE6B8}
\definecolor{projectblue}{HTML}{2563EB}

\title{\textit{Lost on Campus}: Evaluating Embodied Spatial Reasoning of Vision-Language Models in the Wild}

\author{%
  \textbf{Zehan Zheng}$^{1}$\thanks{Equal contribution. \quad $^{\dagger}$Equal advising.} \quad \textbf{Yanyuan Chen}$^{2}$\footnotemark[1] \quad \textbf{Deming Li}$^{1}$ \quad \textbf{Yutao Tang}$^{1}$ \quad \textbf{Jianwen Xie}$^{4}$ \quad \\
  \textbf{Alan Yuille}$^{1}$ \quad \textbf{Jieneng Chen}$^{3}$\footnotemark[2] \quad \textbf{Cheng Peng}$^{2}$\footnotemark[2]\vspace{0.1cm}\\
  \vspace{0.2cm}
  $^{1}$Johns Hopkins University \quad $^{2}$University of Virginia \quad $^{3}$Stanford University \quad  $^{4}$Lambda\\
  {\hypersetup{pdfborder={0 0 0}}\href{https://dyfcalid.github.io/Lost-on-Campus}{\textcolor{projectblue}{\nolinkurl{https://dyfcalid.github.io/Lost-on-Campus}}}}
}

\begin{document}

\maketitle

\begin{abstract}
Embodied Spatial Reasoning (ESR) is central to deploying Vision-Language Models (VLMs) in real-world embodied tasks, yet remains inadequately evaluated. Existing benchmarks primarily assess spatial reasoning in static images or synthetic indoor environments, where semantic shortcuts often suffice, making it difficult to faithfully evaluate spatial reasoning capabilities in the wild. 
In light of this, we present \textit{Lost on Campus}, a benchmark for evaluating ESR in large-scale real-world outdoor 3D environments reconstructed by 3D Gaussian Splatting. 
Our benchmark introduces a unified reasoning-action evaluation framework that seamlessly integrates diagnostic QA for isolated reasoning with closed-loop interactive navigation for active reasoning under multimodal instructions.
To enable fine-grained diagnosis, we evaluate six capability aspects of ESR: action grounding, spatial foresight, metric awareness, goal-directed planning, ego-allocentric localization, and spatio-temporal consistency. 
Extensive experiments reveal that: (i) Compared to indoor settings, real-world outdoor environments impose substantially higher demands on spatial reasoning, where existing models exhibit poor performance; (ii) VLMs still struggle with fine-grained visual alignment, with failures in spatial foresight and ego-allocentric localization emerging as primary bottlenecks; (iii) Improving multimodal interaction capabilities and long-term spatial reasoning is crucial for advancing embodied intelligence. 
These findings highlight that faithful evaluation of ESR demands benchmarks tightly coupling perception, reasoning and action within realistic and continuous environments.
\end{abstract}

\begin{figure}[ht]
  \centering
  \includegraphics[width=0.95\textwidth]{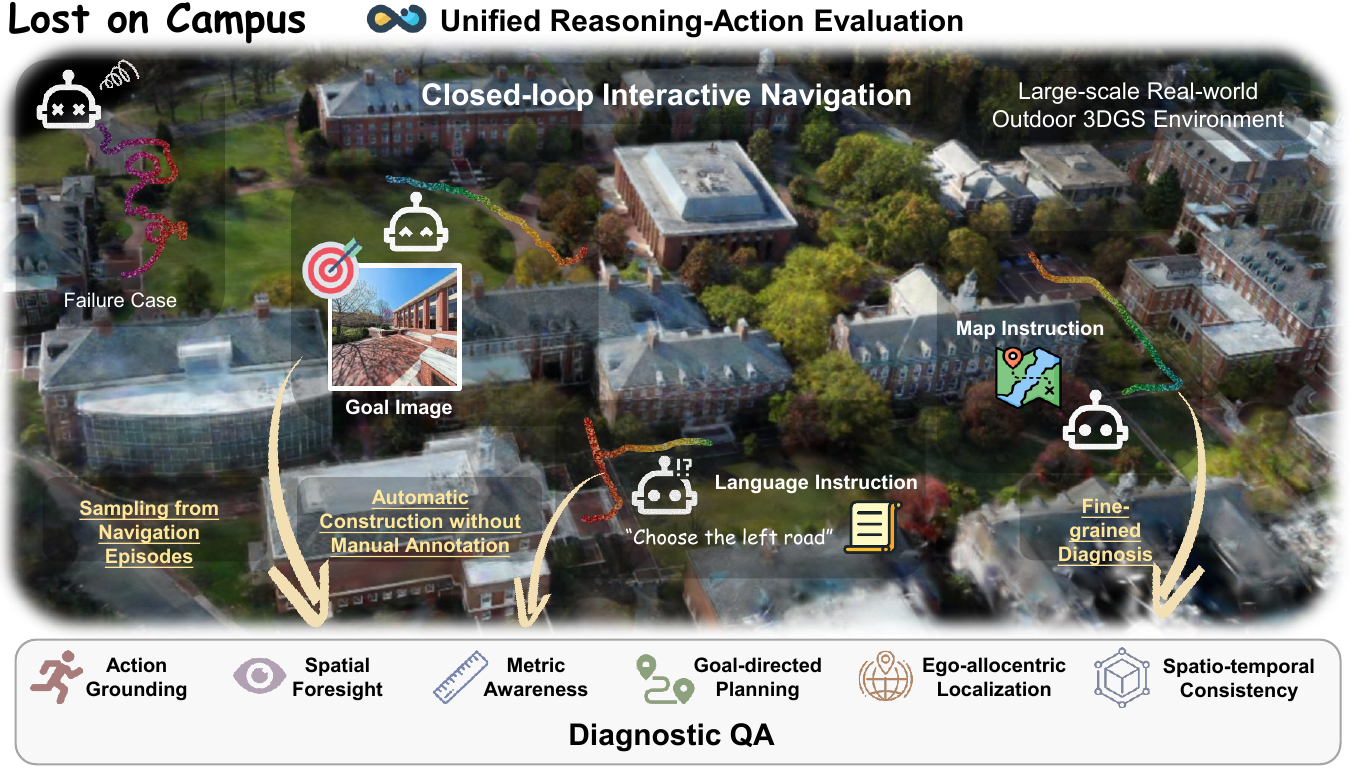}
  \caption{\textbf{Overview of the \textit{Lost on Campus} benchmark.} The evaluation is built upon large-scale photorealistic outdoor 3DGS environments reconstructed from real-world campus captures. This benchmark leverages closed-loop interactive navigation using different spatial instructions, integrating spatial understanding with action execution. Through navigation-grounded QA in an embodied context, it provides a fine-grained probe of six diagnostic ESR capabilities.}
  \label{fig:framework}
\end{figure}

\section{Introduction}
\label{sec:intro}

Spatial reasoning is a foundational capability for Vision-Language Models (VLMs)~\cite{alayrac2022flamingo, anthropic2024claude, bai2025qwen25vl, chen2024internvl, guo2025seed15vl, hurst2024gpt4o, li2024llavaonevision, lu2024ovis, radford2021learning, team2023gemini, qwenteam2025qwen3vl, wang2025internvl35, zhang2024llavanext} that aspire to perceive, navigate, and interact with the physical world.
While recent VLMs have demonstrated remarkable proficiency across a spectrum of visual understanding tasks, a critical question remains: how well can they truly reason about space, particularly in embodied settings where reasoning must translate into action?

Prior works have examined complementary slices of spatial reasoning, yet primarily confine their scope to static images or passive scenes~\cite{xu2025spatialbench,yang2025thinking}.
Recent benchmarks have evolved to incorporate multi-view images to capture viewpoint changes~\cite{gholami2025spatial,wu2025indoor}, or video inputs that evaluate spatio-temporal understanding capabilities~\cite{li2025sti}.
While these efforts advance the evaluation of spatial relationships, they predominantly treat the model as a passive observer. In other words, they assess whether a model can \emph{comprehend} a spatial problem from isolated visual observations, leaving its ability to \emph{act} spatially within a continuous environment largely untested.

Embodied benchmarks and their associated simulators provide interactive environments to address this problem, but they concentrate overwhelmingly on indoor scenes (\textit{e.g.}, Habitat~\cite{savva2019habitat}, AI2-THOR~\cite{kolve2017ai2}).
However, indoor settings inherently exhibit high semantic density within spatially constrained layouts.
In such environments, models can often avoid authentic spatial reasoning by exploiting structural priors or semantic shortcuts, relying on conceptual associations and semantic pattern matching (\textit{e.g.}, locating a refrigerator implies proximity to a kitchen) rather than actual geometric understanding.
While semantically sparse outdoor scenes help decouple the evaluation of spatial reasoning, existing benchmarks still lack realism or continuity. For instance, synthetic platforms like EmbodiedCity~\cite{gao2024embodiedcity} fail to capture real-world fidelity, while street-view datasets such as Touchdown~\cite{chen2019touchdown} are restricted to discrete topological graphs that preclude continuous actions and fine-grained geometric reasoning.
Consequently, a unified evaluation that tightly couples egocentric spatial perception with \textit{embodied interaction} in realistic and continuous environments remains absent.

To fill this gap, we present \textit{Lost on Campus}, a benchmark for evaluating ESR in large-scale real-world outdoor 3D environments. As shown in Figure~\ref{fig:framework}, we reconstruct diverse regions of a university campus using Octree-based Gaussian Splatting~\cite{kerbl20233dgs,ren2025octree}, yielding a photorealistic, view-consistent and continuously navigable simulation platform. It supports long-horizon closed-loop interaction with high-fidelity egocentric observations
for embodied evaluation in real-world open spaces.

Built on this environment, we introduce a unified reasoning-action evaluation framework for VLMs, consisting of two complementary paradigms. 
First, \emph{Closed-Loop Interactive Navigation} evaluates reasoning in action, aiming to assess whether VLMs can effectively translate spatial observations and reasoning into coherent embodied behavior. We further evaluate navigation under three distinct forms of spatial instruction to probe VLMs' spatial reasoning ability in action.
Second, \emph{diagnostic QA} evaluates spatial reasoning in isolation through six diagnostic tasks targeting different aspects of ESR.
Unlike traditional spatial QA on static and disconnected images, our diagnostic protocol is automatically generated from the navigation rollouts across models. Grounded in embodied context, it maps end-to-end task execution back to underlying ESR capabilities, enabling fine-grained evaluation.

We summarize our contributions as follows.
\begin{itemize}

    \item \textbf{Outdoor real-world evaluation environment.}
    We construct a photorealistic and continuous outdoor 3DGS campus environment for embodied VLM evaluation, supporting high-fidelity rendering and long-horizon interactive navigation.

    \item \textbf{Unified reasoning-action evaluation framework.}
    We evaluate VLMs through closed-loop interactive navigation and diagnostic QA, combining reasoning in isolation with reasoning in action for fine-grained analysis of ESR capabilities.

    \item \textbf{Empirical findings on outdoor ESR.}
    We comprehensively benchmark representative open- and closed-source VLMs, show that outdoor ESR remains under-solved, identify spatial foresight and ego-allocentric localization as primary bottlenecks for embodied navigation.

\end{itemize}

\begin{table}[t]
\centering
\caption{Comparison of existing benchmarks related to spatial reasoning. \LEFTcircle\ denotes partial coverage: either disembodied spatial comprehension or limited embodied spatial assessment.}
\label{tab:comparison}
\resizebox{\linewidth}{!}{
\begin{tabular}{lcccccc}
\toprule
\textbf{Benchmark} & \textbf{Scene Source} & \textbf{Env} & \textbf{Motion} & \textbf{Embodied} & \textbf{Spatial Evaluation} & \textbf{3D GT Source} \\
\midrule
SpatialBench~\cite{xu2025spatialbench}        & Image        & Mixed                & N/A        & \xmark & \LEFTcircle  & Monocular Depth Estimation            \\
VSI-Bench~\cite{yang2025thinking}           & Video        & Indoor               & N/A        & \xmark & \LEFTcircle  & Off-the-shelf Dataset           \\
STI-Bench~\cite{li2025sti}           & Video        & Indoor\&Autonomous   & N/A        & \xmark & \LEFTcircle  & Off-the-shelf Dataset   \\
Ego3D-Bench~\cite{gholami2025spatial}         & Video        & Autonomous    & N/A        & \xmark & \LEFTcircle  & Off-the-shelf Dataset          \\
OpenBench~\cite{wu2025indoor}           & Video        & \textbf{Outdoor}     & N/A        & \xmark & \LEFTcircle  & Stereo, LiDAR\&IMU \\
\midrule
REVERIE~\cite{qi2020reverie}             & MP3D         & Indoor               & Discrete   & \cmark & \xmark       & Mesh      \\
NavBench~\cite{qiao2026navbench}           & MP3D         & Indoor               & Discrete   & \cmark & \LEFTcircle  & Mesh      \\
OVON~\cite{yokoyama2024hm3d}                & HM3D         & Indoor               & Continuous & \cmark & \xmark       & Mesh      \\
GOAT-Bench~\cite{khanna2024goat}          & HM3D         & Indoor               & Continuous & \cmark & \xmark       & Mesh      \\
LHPR-VLN~\cite{song2025towards}            & HM3D         & Indoor               & Continuous & \cmark & \xmark       & Mesh      \\
OctoNav~\cite{gao2025octonav}             & HM3D + MP3D  & Indoor               & Continuous & \cmark & \xmark       & Mesh      \\
NavSpace~\cite{yang2025navspace}            & HM3D         & Indoor               & Continuous & \cmark & \LEFTcircle  & Mesh      \\
Sage-Bench~\cite{miao2025towards}          & InteriorGS   & Indoor               & Continuous & \cmark & \xmark       & 3DGS      \\
EmbodiedCity~\cite{gao2024embodiedcity}        & Synthetic    & \textbf{Outdoor}     & Continuous & \cmark & \xmark       & Synthetic \\
Touchdown~\cite{chen2019touchdown}           & Street View  & \textbf{Outdoor}     & Discrete   & \cmark & \xmark       & GPS       \\
\midrule
\textit{\textbf{Lost on Campus}}     & \textbf{Campus 3DGS}  & \textbf{Outdoor}     & \textbf{Continuous} & \cmark & \cmark       & \textbf{GPS, RTK, 3DGS}      \\
\bottomrule
\end{tabular}
}
\end{table}

\section{Related Work}
\label{sec:related}
\paragraph{Spatial reasoning evaluation for VLMs.}
Early evaluations of VLM spatial reasoning focused on reasoning within static single-image scenes, bounding the assessment to qualitative relational understanding~\cite{dongfang2025are, johnson2017clevr, liu2025can, ramakrishnan2025does, shiri2024empirical, stogiannidis2025mind, tang2024sparkle, wang2025spatial457, yao2025lens}. Recent benchmarks have extended spatial evaluation with multi-view images, videos, and 3D annotations~\cite{gholami2025spatial, yang2025mmsi, yeh2026seeing,gong2025space, yang2025thinking, zhang2025do, zhou2025vlm4d}. These efforts introduce richer spatial cues, including viewpoint changes, temporal continuity, depth and 3D geometry across indoor scenes, autonomous-driving scenarios, object-centric 3D datasets and outdoor stereo videos. Such designs substantially broaden the scope of VLM spatial evaluation beyond single images~\cite{gholami2025spatial,li2025sti}. However, they still largely evaluate models as passive observers over pre-collected visual inputs or annotated geometric cues~\cite{yang2025thinking,gholami2025spatial}. The model is asked to answer spatial questions, but its judgments are not coupled with embodied decisions or evaluated through their consequences in navigation. In contrast, our trajectory-grounded QA acts as a diagnostic probe generated from navigation trajectories in a large-scale outdoor environment. It situates spatial understanding within closed-loop embodied behavior.

\paragraph{Embodied navigation benchmarks.}
Planning and navigating through unknown environments remain fundamental challenges for embodied agents. Recent benchmarks evaluate VLMs as high-level decision makers tasked with interpreting egocentric observations and following instructions~\cite{gao2025octonav,miao2025towards}. As summarized in Table~\ref{tab:comparison}, most existing efforts~\cite{khanna2024goat, song2025towards,miao2025towards,qiao2026navbench,yang2025navspace,gao2024embodiedcity} primarily focus on increasing instruction complexity. Although these tasks have driven progress in instruction-following, they concentrate on benchmarking the multimodal understanding of language and visual semantics, without systematically assessing spatial perception and reasoning~\cite{gao2025octonav,miao2025towards}.
Several recent efforts have advanced toward spatial evaluation. For instance, NavBench~\cite{qiao2026navbench} assesses action consequences in spatial understanding; however, it operates within discrete environments, precluding the evaluation of fine-grained egocentric control. Other efforts, such as NavSpace~\cite{yang2025navspace}, advance toward quantitative spatial evaluation in indoor scenes. However, such scenes are limited in spatial scale, semantic diversity and visual variability, hindering evaluation from accurately reflecting real-world conditions.

\paragraph{3D reconstruction and image-based rendering for embodied simulation.}
Simulation environments are central to embodied AI, supporting evaluation and scalable training.
Indoor environment datasets and simulators~\cite{chang2017matterport3d,savva2019habitat} have provided important foundations for embodied reasoning benchmarks.
However, many existing environments are primarily built upon mesh-based scene representations and may lack the visual fidelity needed for fine-grained perception and motion cues.
Meanwhile, existing outdoor environments either rely on discrete panorama graphs~\cite{chen2019touchdown} or synthetic rendering engines~\cite{gao2024embodiedcity}, leaving realistic and continuous outdoor embodied evaluation less explored.
Recent progress in large-scale 3D reconstruction and novel view synthesis, particularly 3DGS~\cite{kerbl20233dgs}, introduces promising solutions to create photorealistic replicas from video captures.
However, large-scale reconstructed environments that combine photorealistic appearance, metric consistency, and continuous outdoor interaction remain limited.
Such environments can complement existing indoor and synthetic or graph-based outdoor platforms, and support broader evaluation of VLMs under realistic egocentric perception and embodied interaction.

\begin{figure}[t]
  \centering
  \includegraphics[width=\textwidth]{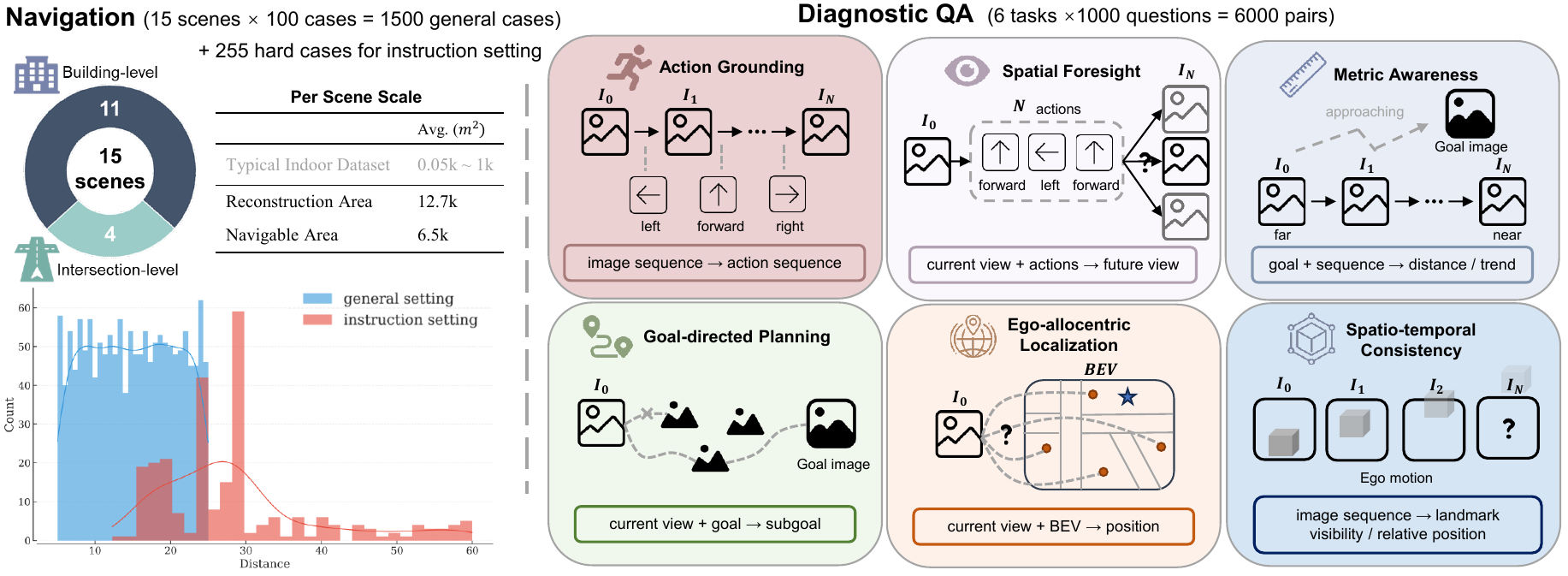}
  \caption{\textbf{Benchmark statistics and diagnostic QA task illustration.} \textit{Left}: Our benchmark covers scenarios of buildings and intersections, achieving a large-scale average navigable area. We further divide cases into general settings involving short-to-medium distances and complex scenarios involving long-distance navigation with instructions. \textit{Right}: Schematic illustrations of the six diagnostic QA tasks, each targeting a fundamental ESR capability (see Section~\ref{sec:diagnostic_qa} for definitions).}
  \label{fig:statistics}
\end{figure}

\section{The \textit{Lost on Campus} Benchmark}

\subsection{The 3D Outdoor Environment}
\label{sec:environment}

\textbf{Data Collection.} The \textit{Lost on Campus} environment is constructed from real-world, large-scale campus captures using Insta360 X4 cameras. The collected panorama frames are split into four perspective images, each with a \ang{120} field of view (FoV). These four frames together cover the horizontal \ang{360} FoV around the camera. 
Overall, this dataset contains \textit{over 50k calibrated video frames} on fifteen academic halls and road intersections, covering an area of approximately $2 \times 10^5~\mathrm{m}^2$.

While frames within a single video are easier to calibrate, frames between videos introduce challenges, \textit{e.g.}, through inconsistent lighting, visual ambiguities, etc. To achieve accurate metric-scale camera calibration across 50k video frames, we use a mobile RTK device to capture centimeter-accurate GPS coordinates for each image. This allows us to (1) perform dense 3D reconstruction on regions where multiple videos overlap, and (2) introduce consistent metric-scale reconstruction, which is strictly required for evaluating VLMs on distance-aware spatial understanding.

\textbf{Photorealistic 3D Reconstruction.} For photorealistic 3D reconstruction and rendering, we employ a modified Octree-GS~\cite{ren2025octree} with (1) appearance and transient embeddings to account for lighting and transient variations and (2) a triangle primitive~\cite{held2025triangle} for better geometry recovery and collision detection. To avoid sky floaters, the sky is independently modeled with a neural field through sky-masking. 
Extended reconstruction details are provided in Appendix~\ref{app:environment}. We further analyze reconstruction fidelity and the potential effect of rendering artifacts on downstream benchmark evaluation in Appendix~\ref{append:reconstruction_analysis}.

\textbf{Environment Characteristics.} As illustrated in Figure~\ref{fig:statistics}, the constructed environment comprises 15 distinct outdoor scenes, categorized into two tiers of topological complexity. \textit{Intersection-level environments} (4 scenes) are reconstructed from three-way or four-way junctions, focusing on evaluating a VLM's spatial decision-making. \textit{Building-level environments} (11 scenes) are reconstructed from images surrounding a building. Quantitatively, each scene features an average 3D reconstructed area of approximately 12.7k m$^2$, with a navigable area of around 6.5k m$^2$.

\subsection{Benchmark Construction}
\label{sec:framework}
\paragraph{Overview.} 
As shown in Figure~\ref{fig:framework},
\textit{Lost on Campus} evaluates embodied spatial reasoning along two complementary axes: closed-loop interactive navigation and diagnostic spatial QA.
For navigation, we formulate all evaluations as image-goal navigation, where an agent must navigate through the 3DGS environment from a start pose to a goal depicted by an image.
For diagnostic QA, we construct a navigation-grounded QA protocol where questions are generated from the agent's completed navigation trajectories within our 3DGS environment, grounding spatial reasoning evaluation in embodied experience.
This automated annotation-free design enables scalable evaluation across models and allows direct investigation of the relationship between navigation performance and underlying spatial understanding.

\paragraph{Navigation case generation.}
\label{sec:nav_settings}
We construct navigation cases through an automated pipeline that samples start--goal camera pose pairs from the raw reconstruction trajectories. In the \textbf{general setting}, we generate 100 cases per scene (1,500 cases in total across 15 scenes), with start--goal distances uniformly distributed in the range of 5--25\,m. This setting provides a standardized evaluation of image-goal navigation. To further investigate the spatial reasoning capabilities of VLMs in complex scenarios and under richer spatial cues, we construct an additional \textbf{instruction setting} comprising 255 challenging cases, including complex scenarios such as long-distance navigation and intersections, to evaluate how VLMs integrate various forms of spatial information into embodied actions.

\paragraph{Diagnostic spatial QA.}
\label{sec:diagnostic_qa}
Beyond end-to-end navigation, we design a fine-grained diagnostic QA protocol comprising six tasks that probe different aspects of ESR.
The QA data is automatically constructed from the navigation outputs of different VLM models.
Each task contains 1,000 questions, yielding a unified evaluation set of 6,000 QA pairs in total.
The six diagnostic tasks, illustrated in Figure~\ref{fig:statistics}, are defined as follows.

\begin{enumerate}
    \item \textbf{Action Grounding.}
    Given a sequence of $N$ consecutive egocentric observation images, the VLM is asked to output the $N{-}1$ primitive actions that produced the observed transitions.
    This task evaluates whether the model can infer ego-motion from visual changes.

    \item \textbf{Spatial Foresight.}
    Given the current observation image and a sequence of primitive actions, the VLM must select the correct post-action observation from a set of candidate images.
    This task probes whether the model can mentally simulate the visual consequence of a given action sequence, demonstrating implicit world modeling capabilities.

    \item \textbf{Metric Awareness.}
    Given a goal image and a sequence of egocentric observations captured at varying distances to the goal, the VLM is asked to identify the closest or farthest frame, or to judge the overall movement trend (approaching vs.\ receding).
    This task evaluates the model's sensitivity to metric distance changes from visual cues.

    \item \textbf{Goal-directed Planning.}
    Given the current observation and a goal image, the VLM must select the correct intermediate subgoal image from candidates, corresponding to a viewpoint that lies along the path to the goal.
    This task assesses the model's ability to decompose a navigation objective into spatial sub-plans.

    \item \textbf{Ego-allocentric Localization.}
    Given the current egocentric observation and a bird's-eye-view (BEV) map, the VLM must identify its own position on the BEV map.
    Alternatively, given a goal image and its own marked position on the BEV, the model needs to localize the goal on the map. 
    This task evaluates perspective-to-allocentric spatial transformation and self-localization.

    \item \textbf{Spatio-temporal Consistency.}
    Given a sequence of egocentric observation images with a specific object highlighted in the first frame, the VLM is asked to determine whether the object remains visible in the final frame, and the object's movement trend or spatial position relative to the agent after the full motion sequence, even when the object is no longer visible and the model must reason about its position from memory. This task probes the model's ability to maintain a coherent spatial representation of tracked objects under ego-motion.
\end{enumerate}

\section{Experiments and Analysis}
\label{sec:experiments}
We evaluate a set of proprietary and open-source foundation models: (1) Proprietary models include Gemini-3-Flash~\cite{google2025gemini3flash} and GPT-5-Mini~\cite{singh2025gpt5}. (2) Open-source models include Qwen3.5-9B/27B~\cite{qwen2026qwen35}, Qwen3-VL-8B/32B-Instruct~\cite{bai2025qwen3vl}, Gemma-4-31B~\cite{team2026gemma4} and InternVL3.5-8B~\cite{wang2025internvl35}. 

\textbf{Implementation Details.} 
The reconstructed 3DGS environments support continuous camera motion, allowing observations to be rendered from arbitrary camera poses. The agent state is represented by a continuous position and orientation, with each primitive action updating the camera pose by the corresponding metric translation or rotation before rendering the next observation. For evaluation, we control the agent through a discrete set of metric action primitives, where each action applies a fixed translation or rotation to the current camera pose. 
At each navigation step, the agent receives RGB observations from three egocentric directions: front, left ($+90^\circ$), and right ($-90^\circ$).
The resolution of each visual observation is set to $512 \times 512$. The action space consists of a set of primitive actions, including moving forward or backward by 1\,m, turning left or right by $45^\circ$, turning left or right by $15^\circ$, and stopping. 
For efficiency, the agent is allowed to predict an action sequence containing multiple primitive actions at each decision step. An episode terminates when the agent outputs the stop action, and the maximum episode length is set to 100 steps. Additional implementation details are provided in Appendix~\ref{append:evaluation_setup}.

\textbf{Metrics.}
For multiple-choice spatial QA, we use \textit{Accuracy} as the primary metric.
For navigation tasks, we report \textit{Success Rate} (\textit{SR}), \textit{Oracle Success Rate} (\textit{OSR}), and \textit{Success weighted by Path Length} (\textit{SPL}) under both 2.5\,m and 5.0\,m success thresholds.
\textit{SR} measures whether the agent stops within the goal region, \textit{OSR} measures whether the agent ever reaches the goal region during the trajectory, and \textit{SPL} accounts for path efficiency among successful episodes.
We further report auxiliary navigation metrics averaged over trajectories: \textit{Distance to Goal} (\textit{DTG}), final-view similarity (\textit{Sim.}), progress score (\textit{Prog.}), trajectory length (\textit{Traj.}), inference requests (\textit{Calls}), and primitive actions (\textit{Actions}).
\textit{DTG} measures the final distance from the agent to the goal after termination.
\textit{Sim.} measures the image similarity using DINOv2~\cite{oquab2023dinov2} between the agent's final front view and the goal image. \textit{Prog.} measures the average effective distance gained per step toward the goal.
\textit{Traj.} records the total traveled distance.
\textit{Calls} counts the number of inference requests.
\textit{Actions} counts the total number of executed primitive actions, where each inference request may return multiple primitive actions.

\subsection{Main Navigation Results}

\begin{table*}[t]
    \centering
    \caption{Main navigation results on \textit{Lost on Campus}. All metrics are reported on the general setting of 1500 image-goal cases. \textbf{Bold} and \underline{underline} denote the best and second-best results, respectively.}
    \label{tab:nav_main}
    \vspace{0.1cm}
    \definecolor{groupblue}{HTML}{E2E8F0}
    \definecolor{groupgray}{HTML}{F1F5F9}
    
    \resizebox{\textwidth}{!}{%
        \renewcommand{\arraystretch}{1.5}
        \begin{tabular}{l *{12}{c}}
            \toprule[1.5pt]
            \textbf{Model} & \textbf{SR@2.5} & \textbf{SR@5} & \textbf{OSR@2.5} & \textbf{OSR@5} & \textbf{SPL@2.5} & \textbf{SPL@5} & \textbf{DTG} & \textbf{Sim.} & \textbf{Prog.} & \textbf{Traj.} & \textbf{Calls} & \textbf{Actions} \\
            \midrule
            
            \rowcolor{groupblue}
            \multicolumn{13}{l}{\textit{\textbf{Proprietary Models}}} \\
            Gemini-3-Flash & \textbf{25.73} & \textbf{49.60} & \textbf{36.13} & \textbf{67.73} & \textbf{17.29} & \underline{31.16} & \underline{10.11} & \textbf{0.699} & 0.27 & 39.14 & \underline{21.7} & 66.2 \\
            GPT-5-Mini     & \underline{11.80} & 30.60 & 21.47 & \underline{49.60} & \underline{9.54}  & 24.34 & 13.81 & 0.615 & 0.22 & 35.45 & 24.4 & 67.5 \\
            \midrule
            
            \rowcolor{groupgray}
            \multicolumn{13}{l}{\textit{\textbf{Open-source Models} ($\sim$30B)}} \\
            Qwen3.5-27B   & 9.93  & 34.07 & 11.33 & 36.47 & 9.05  & \textbf{31.91} & \textbf{9.48}  & \textbf{0.699} & \textbf{0.51} & \textbf{14.49} & 28.7 & 44.6 \\
            Qwen3-VL-32B-Instruct     & 6.33  & 28.07 & 8.33  & 32.80 & 5.49  & 24.99 & 11.30  & 0.662 & \underline{0.42} & \underline{20.27} & 22.2 & \underline{40.3} \\
            Gemma-4-31B    & 10.67 & \underline{36.20} & 13.47 & 40.67 & 8.08  & 28.05 & 11.21 & \underline{0.696} & 0.35 & 27.41 & 24.3 & 54.3 \\
            \midrule
            
            \rowcolor{groupgray}
            \multicolumn{13}{l}{\textit{\textbf{Open-source Models} ($<$10B)}} \\
            Qwen3.5-9B    & 6.33  & 20.47 & 12.87 & 30.00 & 4.73  & 15.49 & 13.00    & 0.571 & 0.19 & 23.29 & \textbf{15.0} & \textbf{36.4} \\
            Qwen3-VL-8B-Instruct      & 10.67 & 25.53 & 17.27 & 39.13 & 8.58  & 20.78 & 13.81 & 0.606 & 0.20 & 27.12 & 40.9 & 60.2 \\
            InternVL3.5-8B& 5.33  & 13.73 & \underline{29.60} & 48.60 & 3.38  & 8.08  & 17.12 & 0.528 & 0.01 & 44.14 & 74.1 & 86.0 \\
            
            \bottomrule[1.5pt]
        \end{tabular}%
    }
\end{table*}

\begin{table}[t]
\small
\centering
\caption{\textbf{Effect of post-hoc pose optimization.} On the basis of Table~\ref{tab:nav_main}, we further apply 3DGS-based pose optimization after the VLM agent stops and does not change the executed action sequence. The $\Delta$ metrics measure the contribution of pose/localization errors to evaluation.}
\label{tab:pose_opt}
\renewcommand{\arraystretch}{1.25}
\resizebox{0.6\textwidth}{!}{
\begin{tabular}{lcccc}
\toprule
 Model & $\Delta$SR@2.5$\uparrow$ & $\Delta$SR@5$\uparrow$ & $\Delta$DTG$\downarrow$ & $\Delta$Sim$\uparrow$ \\
\midrule
Gemini-3-Flash & +10.00 & +2.33 & -0.53 & +0.026 \\
GPT-5-Mini & +6.40 & +3.80 & -0.34 & +0.012 \\
Qwen3.5-27B & +8.60 & +5.60 & -0.46 & +0.015 \\
Qwen3-VL-32B-Instruct & +7.14 & +5.66 & -0.41 & +0.011 \\
Gemma-4-31B & +10.00 & +5.13 & -0.47 & +0.014 \\
Qwen3.5-9B & +3.47 & +2.73 & -0.19 & +0.005 \\
Qwen3-VL-8B-Instruct & +5.33 & +2.74 & -0.30 & +0.014 \\
InternVL3.5-8B & +1.94 & +0.60 & -0.09 & +0.001 \\

\bottomrule
\end{tabular}
}
\end{table}

We first evaluate VLMs on the 1,500 general-setting cases for image-goal navigation. 
Table~\ref{tab:nav_main} reports the standard navigation metrics, and Table~\ref{tab:pose_opt} isolates the contribution of fine-grained pose alignment. 
Additional comparisons with specialized embodied and navigation models are provided in Appendix~\ref{append:specialized_baselines}.

\textbf{Outdoor ESR remains severely under-solved.}
Even the strongest model, Gemini-3-Flash, attains only 49.60\% SR@5 and 25.73\% SR@2.5, while no open-source model exceeds 11\% SR@2.5.
This is far below typical indoor image-goal navigation results, indicating that real-world outdoor embodied spatial reasoning is substantially harder than static image-level spatial understanding, in part because outdoor scenes lack the strong indoor semantic shortcuts on which VLMs often rely.

\textbf{Models often approach the goal but fail to localize and stop correctly.}
Across nearly all models, we observe a large gap between SR@2.5 and SR@5 and also a wide gap between OSR and SR: the agent frequently passes near the target but fails to terminate at the correct location.
This is most extreme for InternVL3.5-8B, which attains a relatively high OSR@5 of 48.60\% yet collapses to only 5.33\% SR@2.5.
Applying post-hoc 3DGS pose optimization at the agent's final stop (Table~\ref{tab:pose_opt}) recovers $+1.94$ to $+10.00$ SR@2.5 across models, confirming that endpoint localization, goal verification, and self-aware stopping are dominant failure modes.

\textbf{Progress and efficiency are not sufficient for success.}
Strong intermediate behavior does not guarantee successful termination.
Qwen3.5-27B records the highest SPL@5 (31.91\%, exceeding Gemini-3-Flash's 31.16\%), the lowest DTG (9.48\,m), and the highest progress score (Prog $=$ 0.51), yet its SR@2.5 (9.93\%) trails Gemini-3-Flash by more than 15 points.
This shows that efficient movement must be coupled with reliable goal recognition and spatial state estimation; otherwise, agents that traverse the environment efficiently still fail to commit to the correct stop.

\newcommand{\rot}[1]{\makebox[1.5em][l]{\rotatebox[origin=bl]{30}{\makecell[l]{#1}}}}

\begin{table*}[t]
    \centering
    \small
    \caption{Diagnostic spatial QA results on \textit{Lost on Campus}. Left table shows the Accuracy (\%) for each ESR capability of corresponding task, and the right radar chart provides a comprehensive comparison.}
    \label{tab:qa}
    \begin{minipage}[c]{0.6\linewidth}
        \definecolor{groupblue}{HTML}{E2E8F0}
        \definecolor{groupgray}{HTML}{F1F5F9}
        \resizebox{\textwidth}{!}{
            \renewcommand{\arraystretch}{1.18}
            \begin{tabular}{l *{7}{c}}
                \textbf{Model} & 
                
                \rot{\scriptsize{Action}} & 
                \rot{\scriptsize{Foresight}} & 
                \rot{\scriptsize{Metric}} & 
                \rot{\scriptsize{Planning}} & 
                \rot{\scriptsize{Localize}} & 
                \rot{\scriptsize{Spatio-temp.}} & 
                Avg. \\
                \midrule

                \rowcolor{groupblue}
                \multicolumn{8}{l}{\textit{\textbf{Baseline}}} \\
                Chance & 16.67 & 25.0 & 23.3 & 25.0 & 25.0 & 25.0 & 23.33 \\
                \midrule
                
                \rowcolor{groupblue}
                \multicolumn{8}{l}{\textit{\textbf{Proprietary Models}}} \\
                Gemini-3-Flash & 75.16 & \textbf{61.9} & \textbf{62.2} & \textbf{75.4} & 41.7 & \textbf{62.4} & \textbf{63.13} \\
                GPT-5-Mini     & 58.53 & 53.3  & 50.6  & 61.4 & 40.0 & \underline{55.2}  & 53.17 \\
                \midrule
                
                \rowcolor{groupgray}
                \multicolumn{8}{l}{\textit{\textbf{Open-source Models} ($\sim$30B)}} \\
                Qwen3.5-27B    & \textbf{82.52} & 55.3  & \underline{61.7}  & 72.2  & 43.2  & 50.0    & \underline{60.82} \\
                Qwen3-VL-32B-Instruct      & 65.25  & 52.2  & 44.7  & 65.7  & 29.1  & 53.8  & 51.79 \\
                Gemma-4-31B     & \underline{76.01} & \underline{58.2}  & 59.0  & \underline{72.5}  & \textbf{44.4}  & 54.3  & 60.74 \\

                \midrule
                \rowcolor{groupgray}
                \multicolumn{8}{l}{\textit{\textbf{Open-source Models} ($<$10B)}} \\
                Qwen3.5-9B     & 58.40 & 52.4 & 41.8 & 56.0 & 42.5 & 49.2 & 50.05 \\
                Qwen3-VL-8B-Instruct       & 52.36 & 30.5 & 38.0 & 53.0 & \underline{43.9} & 48.5 & 44.38 \\
                InternVL3.5-8B & 50.91 & 34.7 & 41.2 & 54.7 & 32.2 & 35.5 & 41.54 \\
                
                \bottomrule[1.5pt]
            \end{tabular}%
            }
    \end{minipage}%
    \hfill
    \begin{minipage}[c]{0.4\linewidth}
    \centering
    \includegraphics[width=\linewidth]{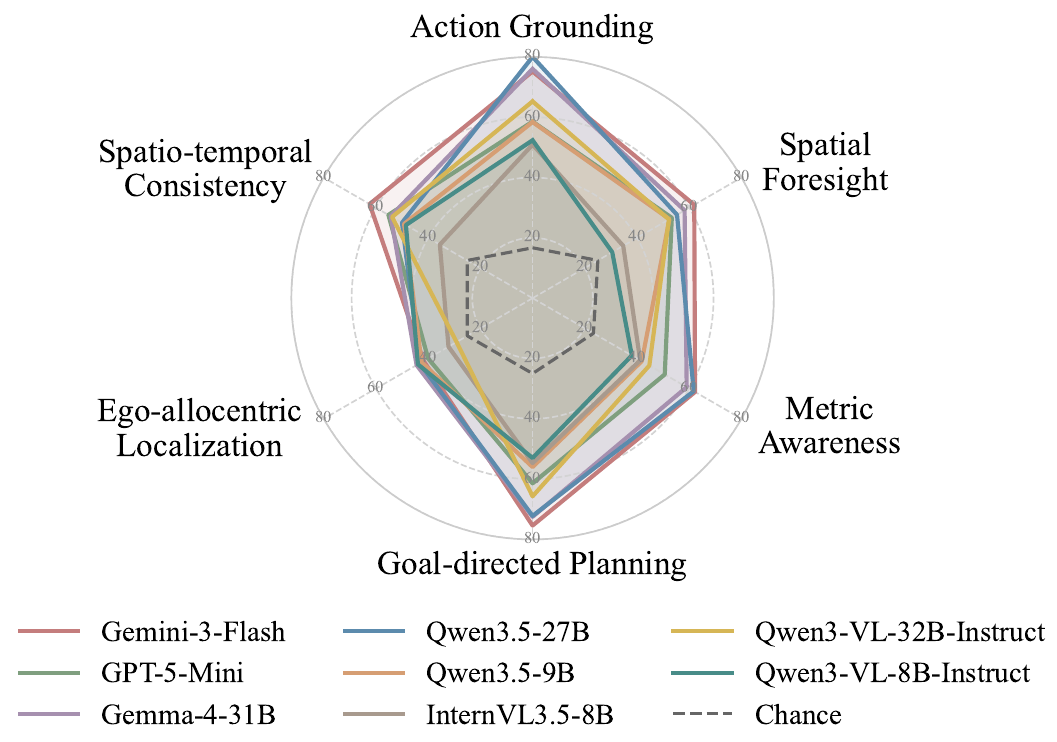}
    \end{minipage}

\end{table*}

\subsection{Diagnostic Capability Breakdown}

We evaluate the same VLMs on the diagnostic spatial QA protocol introduced in Section~\ref{sec:diagnostic_qa}, and use them as complementary diagnostic probes of spatial capabilities involved in the embodied perception--action loop.

\textbf{Spatial capabilities exhibit highly non-uniform performance.}
Averaged across the eight evaluated models, the six tasks form a clear difficulty hierarchy:
Action Grounding (64.9\%) and Goal-directed Planning (63.9\%) are relatively stronger, while Spatial Foresight (49.8\%) and Ego-allocentric Localization (39.6\%, only $\sim$15 points above 4-way chance) remain consistently weak.
Even the strongest model, Gemini-3-Flash, scores only 41.7\% on Ego-allocentric Localization, and Spatial Foresight degrades sharply for smaller models (down to 30.5\% for Qwen3-VL-8B-Instruct).

\subsection{Construct Validity: Diagnostic QA vs. Navigation}
\begin{table}[t]
\centering
\caption{
\textbf{Oracle ablation study for localization information.}
All values are improvements over the original baseline.
}
\label{tab:oracle_localization}
\small
\resizebox{0.95\linewidth}{!}{
\begin{tabular}{lcccccc}
\toprule
\textbf{Setting}
& \textbf{BEV Map}
& \textbf{Goal Pos.}
& \textbf{Agent Pos.}
& $\boldsymbol{\Delta}$\textbf{SR@2.5}
& $\boldsymbol{\Delta}$\textbf{SR@5}
& $\boldsymbol{\Delta}$\textbf{OSR@5} \\
\midrule

(a) Original baseline
& \xmark
& \xmark
& \xmark
& 0.00
& 0.00
& 0.00 \\

(b) BEV-only
& \cmark
& \xmark
& \xmark
& +0.67
& +4.00
& +14.00 \\

(c) Position-only
& \cmark
& \xmark
& \cmark
& +1.34
& +4.00
& +11.34 \\

(d) Goal-only
& \cmark
& \cmark
& \xmark
& +1.34
& -2.67
& +4.00 \\

(e) Full Oracle
& \cmark
& \cmark
& \cmark
& \textbf{+11.34}
& \textbf{+10.00}
& \textbf{+20.67} \\

\bottomrule
\end{tabular}
}
\end{table}

\begin{table}[t]
\centering
\caption{
\textbf{Oracle ablation study for spatial foresight.}
All values are improvements over the original baseline.
}
\label{tab:oracle_foresight}
\small
\resizebox{0.78\linewidth}{!}{
\begin{tabular}{lccc}
\toprule
\textbf{Setting}
& $\boldsymbol{\Delta}$\textbf{SR@2.5}
& $\boldsymbol{\Delta}$\textbf{SR@5}
& $\boldsymbol{\Delta}$\textbf{OSR@5} \\
\midrule

(a) Original baseline
& 0.00
& 0.00
& 0.00 \\

(b) Oracle Foresight w/ DINOv2
& +6.00
& +1.33
& +5.33 \\

(c) Oracle Foresight w/ VLM
& +10.00
& +4.67
& +12.67 \\

(d) Oracle Foresight w/ shortest-DTG
& \textbf{+26.00}
& \textbf{+38.67}
& \textbf{+33.33} \\

\bottomrule
\end{tabular}
}
\end{table}

We examine the joint behavior of isolated reasoning (Table~\ref{tab:qa}) and end-to-end navigation (Table~\ref{tab:nav_main}) to test the construct validity of our framework.

\textbf{Diagnostic QA is broadly aligned with navigation performance.}
At the aggregate level, model rankings by average QA accuracy broadly track navigation performance measured by SR@5. Gemini-3-Flash leads both (63.13\% / 49.60\%), InternVL3.5-8B trails both (41.54\% / 13.73\%), and the two near-equivalent 30B models also achieve nearly the same SR@5, indicating a consistent association between isolated spatial reasoning performance and embodied navigation success.

\textbf{Controlled intervention on localization.}
To examine whether localization-related errors materially affect navigation performance, we conduct controlled oracle interventions on Qwen3.5-27B using a fixed scene-balanced subset of 150 episodes. We progressively introduce spatial information, including the BEV map, oracle goal position, and oracle agent pose (position and heading), as summarized in Table~\ref{tab:oracle_localization}.  Compared with the original baseline, the (b) BEV-only, (c) Position-only, and (d) Goal-only settings show limited SR improvements. In contrast, (e) Full Oracle, which directly provides oracle goal position, agent position, and heading on the BEV, without requiring the model to infer them, yields +11.34 SR@2.5, +10.00 SR@5, and +20.67 OSR@5 over (a) Original baseline.
Moreover, compared with (d) Goal-only, Full Oracle additionally provides the oracle current position and heading, yielding +10.00 SR@2.5, +12.67 SR@5, and +16.67 OSR@5. These gains estimate the potential navigation improvement if the model were able to infer these quantities accurately. Together with the consistently low Ego-allocentric Localization QA scores across models, these results indicate that localization-related errors are an important contributor to navigation failure.

\textbf{Controlled intervention on spatial foresight.}
We further examine whether access to action-conditioned future observations improves navigation performance, using the same Qwen3.5-27B model and the same scene-balanced 150-episode subset. At each decision step, the model proposes multiple candidate action sequences, and the environment renders the true future observation resulting from each candidate. We then select the executed candidate using three strategies: (b) DINOv2 similarity between the rendered future observations and the goal image, (c) the VLM directly inspecting and reasoning over the rendered future observations, and (d) an oracle selecting the candidate with the shortest distance to the goal. The results are summarized in Table~\ref{tab:oracle_foresight}. The results show that navigation performance improves substantially when action-conditioned future observations are provided. Together with the consistently weak Spatial Foresight QA performance, this suggests that foresight-related limitations contribute to closed-loop navigation failures.
We additionally analyze potential shortcuts in both diagnostic QA and
navigation, including linguistic and label-distribution biases as well as
simple exploration and visual-matching baselines; see
Appendix~\ref{append:shortcut_analysis}.

\textbf{Diagnostic QA and navigation therefore provide complementary views of ESR.}
Diagnostic QA isolates capability-specific weaknesses under controlled conditions, whereas closed-loop navigation evaluates whether these capabilities can be integrated consistently over a sequence of perception--action decisions.
The agreement between diagnostic patterns and navigation behavior, together with the controlled interventions above, supports the use of the two evaluation modes as complementary components of the \textit{Lost on Campus} benchmark.

\subsection{Ablation on Navigation Prompt Design}

\begin{table}[t]
\small
\centering
\caption{Ablation study based on Qwen3.5-27B in the general setting of image-goal navigation.}
\label{tab:ablation}
\renewcommand{\arraystretch}{1.18}
\begin{tabular}{lcccc}
\toprule
 Setting & SR@5$\uparrow$ & SPL@5$\uparrow$ & DTG$\downarrow$ & Traj. L$\downarrow$ \\
\midrule
\rowcolor{gray!12}
Baseline & 34.07 & 31.91 & 9.48 & 14.49 \\
w/o Principle & 27.33 & 27.45 & 9.89 & 11.09 \\
w/o Navigation CoT & 31.27 & 28.22 & 9.59 & 15.71 \\
w/o Thought & 31.20 & 21.46 & 14.25 & 48.82 \\

\bottomrule
\end{tabular}
\end{table}

We ablate three components of our navigation prompt template on Qwen3.5-27B: spatial reasoning \textit{Principles}, structured chain-of-thought \textit{Navigation CoT}, and explicit \textit{Thought} output field (Table~\ref{tab:ablation}).

\textbf{All three components contribute, with distinct roles in success and efficiency.}
Removing the \textit{Principles} causes the largest SR drop ($-6.74$ to 27.33\%) and the shortest trajectory (11.09\,m), indicating that without spatial reasoning guidance the agent terminates prematurely.
Removing \textit{Navigation CoT} yields moderate degradation in both SR ($-2.80$) and SPL ($-3.69$), showing that step-level reasoning supports more reliable action selection.
Removing the \textit{Thought} field causes only a small SR drop ($-2.87$) but a catastrophic collapse in efficiency: SPL falls by $10.45$ points, DTG rises to 14.25\,m, and trajectory length explodes to 48.82\,m ($>3\times$ baseline). Without explicit reasoning before action, the agent wanders extensively before occasionally stumbling onto the goal.
Together, these results justify all three components: \textit{Principles} guide goal-directed termination, \textit{Navigation CoT} supports reliable per-step decisions and the \textit{Thought} field is essential for path efficiency.

\subsection{Probing Spatial Reasoning in Embodied Interaction}
\label{sec:nav_eval}
\begin{table*}[t]
\centering
\small
\caption{\textbf{Controlled diagnostic comparison for spatial reasoning in embodied interaction.}
All results are reported under the 5.0\,m success threshold.
\textit{Image-Goal Nav.} reports absolute SR and OSR (\%).
Other rows report changes ($\Delta$) relative to \textit{Image-Goal Nav.} for the same model.
\protect\raisebox{0.4ex}{\protect\colorbox{green!22}{\protect\rule{0.6em}{0em}}} highlights positive changes.
Additional auxiliary metrics are reported in Appendix~\ref{append:full_controlled_results}.}
\label{tab:nav_instruction}
\setlength{\tabcolsep}{5.2pt}
\renewcommand{\arraystretch}{1.18}
\begin{tabular}{@{}l cc cc cc cc@{}}
\toprule
\multirow{2}{*}{\textbf{Setting}}
  & \multicolumn{2}{c}{\textbf{Qwen3-VL-8B}}
  & \multicolumn{2}{c}{\textbf{Qwen3.5-9B}}
  & \multicolumn{2}{c}{\textbf{Qwen3.5-27B}}
  & \multicolumn{2}{c}{\textbf{Qwen3-VL-32B}} \\
\cmidrule(lr){2-3}\cmidrule(lr){4-5}\cmidrule(lr){6-7}\cmidrule(lr){8-9}
  & SR$\uparrow$ & OSR$\uparrow$
  & SR$\uparrow$ & OSR$\uparrow$
  & SR$\uparrow$ & OSR$\uparrow$
  & SR$\uparrow$ & OSR$\uparrow$ \\
\midrule
\rowcolor{gray!12}
\textit{Image-Goal Nav.} (Abs.)
  & 4.71 & 10.20
  & 0.39 & 0.78
  & 3.92 & 4.71
  & 5.88 & 7.84 \\
\addlinespace[2pt]
$\Delta$ \textit{w/ Language Instruction}
  & $-$1.57 & $-$4.71
  & \cellcolor{green!18}+2.35 & \cellcolor{green!16}+1.96
  & \cellcolor{green!16}+1.96 & \cellcolor{green!12}+1.18
  & \cellcolor{green!10}+0.78 & \cellcolor{green!22}+3.14 \\
$\Delta$ \textit{w/ Goal-Annotated BEV}
  & \cellcolor{green!10}+0.78 & $-$1.57
  & \cellcolor{green!8}+0.39 & 0.00
  & \cellcolor{green!30}+4.71 & \cellcolor{green!33}+5.49
  & \cellcolor{green!22}+3.14 & \cellcolor{green!41}+7.06 \\
$\Delta$ \textit{w/ Visual Route Map}
  & $-$2.75 & $-$6.67
  & \cellcolor{green!10}+0.78 & \cellcolor{green!10}+0.78
  & \cellcolor{green!28}+4.31 & \cellcolor{green!32}+5.10
  & \cellcolor{green!16}+1.96 & \cellcolor{green!37}+6.27 \\
\bottomrule
\end{tabular}
\end{table*}

Embodied spatial reasoning requires more than recognizing spatial relations; it requires using spatial information to guide continuous action under evolving egocentric observations. We construct a long-range hard-case subset across all reconstructed outdoor scenes, covering both building-level and intersection-level environments. The dataset statistics are shown in Figure~\ref{fig:statistics}. 
All settings are evaluated on the same navigation cases: the start location, goal location, and initial orientation.
The settings differ in the form and granularity of spatial input provided to the model:

\begin{itemize}
    \item \textbf{Language Instruction:} The model additionally receives egocentric textual route instructions at the beginning of navigation.
    \item \textbf{Goal-Annotated BEV:} Each start--goal pair is augmented with a bird's-eye-view (BEV) image on which only the goal location is marked.
    \item \textbf{Visual Route Map:} Route annotations are overlaid on the BEV image, visualizing the same route-level information as the \textit{Language Instruction} setting.
\end{itemize}

\textbf{Long-range navigation and reasoning remain limited.} We begin with \textit{Image-Goal Nav.}, where the model receives the raw start--goal case and directly explores the environment. The absolute performance is reported in Table~\ref{tab:nav_instruction}. Overall, SR remains below 10\% across all models and settings, indicating that current VLMs remain far from advanced embodied spatial reasoning in real-world outdoor environments. External spatial information brings certain improvements over \textit{Image-Goal Nav.}, demonstrating that current VLMs possess partial spatial reasoning ability during embodied interaction.
However, these gains are limited. In particular, Qwen3-VL-8B-Instruct exhibits degradation under multiple settings, suggesting that smaller multimodal models may fail to integrate external spatial cues into the embodied interaction loop. By contrast, larger models show more consistent gains, indicating a stronger capacity to align spatial representations with egocentric perception. 

\textbf{Structured visual-spatial representations are more effective for larger VLMs than language-only route instructions.}
Among the larger models, the gains from \textit{Language Instruction} are smaller than those from the two structured visual variants.
For Qwen3.5-27B, \textit{Goal-Annotated BEV} and \textit{Visual Route Map} improve SR by $+4.71$ and $+4.31$, respectively, compared with $+1.96$ from \textit{Language Instruction}.
For Qwen3-VL-32B-Instruct, \textit{Goal-Annotated BEV} yields a $+3.14$ SR gain and the largest OSR gain of $+7.06$, while \textit{Language Instruction} improves SR by only $+0.78$.
This suggests that larger VLMs benefit more from explicit visual spatial structure than from language-only route descriptions in embodied spatial reasoning.
We provide more analyses in Appendix~\ref{append:full_controlled_results}.


\section{Conclusion}
\label{sec:conclusion}

We introduced \textit{Lost on Campus}, a large-scale real-world outdoor benchmark for evaluating Embodied Spatial Reasoning in VLMs. By combining a photorealistic and continuously navigable 3DGS campus environment with closed-loop interactive navigation and diagnostic spatial QA, our benchmark evaluates both spatial reasoning in action and fine-grained spatial understanding. Extensive experiments show that current VLMs remain limited in outdoor ESR, with spatial foresight and ego-allocentric localization emerging as key bottlenecks. Results under spatially guided navigation further reveal a persistent gap between spatial instruction and action execution. 

\vspace{0.2cm}

\label{sec:limitations}

\textbf{Limitations}. While \textit{Lost on Campus} advances the evaluation of embodied spatial reasoning in realistic outdoor settings, several limitations remain. 
All scenes are reconstructed from a single university campus using static captures, which may not fully represent the geographic, architectural, and dynamic diversity of real-world outdoor environments (\textit{e.g.}, urban streets, changing weather, moving pedestrians). 
The current action space consists of predefined primitive actions that enable reproducible evaluation but do not capture the continuous motor control required by real robotic platforms. 
Since the environments are reconstructed using 3DGS, they may still exhibit inherent differences from the physical world, including view-synthesis artifacts and missing dynamic elements. We empirically analyze reconstruction fidelity and the effect of rendering artifacts in Appendix~\ref{append:reconstruction_analysis}.
Finally, the diagnostic QA protocol is automatically generated from navigation rollouts, which enables scalability but may introduce biases tied to the specific trajectories used for construction.


\label{sec:broader_impact}

\textbf{Broader Impact}. We view \textit{Lost on Campus} as a positive contribution to AI safety: by providing a rigorous, fine-grained benchmark for outdoor embodied spatial reasoning, it enables the community to precisely identify where current VLMs fall short, such as in spatial foresight and self-localization, before these models are deployed in safety-critical settings like autonomous navigation or assistive robotics. Rather than serving solely as a cautionary tool, the benchmark offers a constructive pathway toward more reliable embodied agents through targeted, capability-level evaluation. We note that all visual data undergo strict anonymization to remove personally identifiable information, and that the large-scale 3DGS reconstruction and VLM evaluation entail substantial computational costs. The benchmark release is restricted to academic research under responsible-use guidelines to mitigate potential misuse.

\begin{ack}
This work was supported by the Office of Naval Research (ONR) under grants N00014-23-1-2641 and National Eye Institute (NEI) with Award ID: R01EY037193. We thank the Johns Hopkins Research IT team (IT@JH) for their support and computing infrastructure.

This research is supported by the Intelligence Advanced Research Projects Activity (IARPA) via Department of Interior/ Interior Business Center (DOI/IBC) contract number 140D0423C0076. The U.S. Government is authorized to reproduce and distribute reprints for Governmental purposes notwithstanding any copyright annotation thereon. The views and conclusions contained herein are those of the authors and should not be interpreted as necessarily representing the official policies or endorsements, either expressed or implied, of IARPA, DOI/IBC, or the U.S. Government.

\end{ack}

\bibliographystyle{plainnat}
{\small
\bibliography{references}

@article{alayrac2022flamingo,
  title={Flamingo: a visual language model for few-shot learning},
  author={Alayrac, Jean-Baptiste and Donahue, Jeff and Luc, Pauline and Miech, Antoine and Barr, Iain and Hasson, Yana and Lenc, Karel and Mensch, Arthur and Millican, Katherine and Reynolds, Malcolm and others},
  journal={Advances in neural information processing systems},
  volume={35},
  pages={23716--23736},
  year={2022}
}

@misc{anthropic2024claude,
  title={Claude},
  author={Anthropic},
  year={2024},
  howpublished={\url{https://www.anthropic.com/claude}}
}

@article{bai2025qwen25vl,
  title={Qwen2.5-vl technical report},
  author={Bai, Shuai and Chen, Keqin and Liu, Xuejing and Wang, Jialin and Ge, Wenbin and Song, Sibo and Dang, Kai and Wang, Peng and Wang, Shijie and Tang, Jun and others},
  journal={arXiv preprint arXiv:2502.13923},
  year={2025}
}

@inproceedings{chen2024internvl,
  title={Internvl: Scaling up vision foundation models and aligning for generic visual-linguistic tasks},
  author={Chen, Zhe and Wu, Jiannan and Wang, Wenhai and Su, Weijie and Chen, Guo and Xing, Sen and Zhong, Muyan and Zhang, Qinglong and Zhu, Xizhou and Lu, Lewei and others},
  booktitle={Proceedings of the IEEE/CVF conference on computer vision and pattern recognition},
  pages={24185--24198},
  year={2024}
}

@misc{guo2025seed15vl,
  title={Seed1.5-vl technical report},
  author={Guo, Dong and Wu, Faming and Zhu, Feida and Leng, Fuxing and Shi, Guang and Chen, Haobin and Fan, Haoqi and Wang, Jian and Jiang, Jianyu and Wang, Jiawei and others},
  journal={arXiv preprint arXiv:2505.07062},
  year={2025}
}

@article{hurst2024gpt4o,
  title={Gpt-4o system card},
  author={Hurst, Aaron and Lerer, Adam and Goucher, Adam P and Perelman, Adam and Ramesh, Aditya and Clark, Aidan and Ostrow, AJ and Welihinda, Akila and Hayes, Alan and Radford, Alec and others},
  journal={arXiv preprint arXiv:2410.21276},
  year={2024}
}

@article{li2024llavaonevision,
  title={Llava-onevision: Easy visual task transfer},
  author={Li, Bo and Zhang, Yuanhan and Guo, Dong and Zhang, Renrui and Li, Feng and Zhang, Hao and Zhang, Kaichen and Zhang, Peiyuan and Li, Yanwei and Liu, Ziwei and others},
  journal={arXiv preprint arXiv:2408.03326},
  year={2024}
}

@article{lu2024ovis,
  title={Ovis: Structural embedding alignment for multimodal large language model},
  author={Lu, Shiyin and Li, Yang and Chen, Qing-Guo and Xu, Zhao and Luo, Weihua and Zhang, Kaifu and Ye, Han-Jia},
  journal={arXiv preprint arXiv:2405.20797},
  year={2024}
}

@inproceedings{radford2021learning,
  title={Learning transferable visual models from natural language supervision},
  author={Radford, Alec and Kim, Jong Wook and Hallacy, Chris and Ramesh, Aditya and Goh, Gabriel and Agarwal, Sandhini and Sastry, Girish and Askell, Amanda and Mishkin, Pamela and Clark, Jack and others},
  booktitle={International conference on machine learning},
  pages={8748--8763},
  year={2021},
  organization={PmLR}
}

@article{team2023gemini,
  title={Gemini: a family of highly capable multimodal models},
  author={Team, Gemini and Anil, Rohan and Borgeaud, Sebastian and Alayrac, Jean-Baptiste and Yu, Jiahui and Soricut, Radu and Schalkwyk, Johan and Dai, Andrew M and Hauth, Anja and Millican, Katie and others},
  journal={arXiv preprint arXiv:2312.11805},
  year={2023}
}

@misc{qwenteam2025qwen3vl,
  title={Qwen3-vl: The multimodal large language model series},
  author={Qwen Team},
  year={2025},
  howpublished={\url{https://github.com/QwenLM/Qwen3-VL}}
}

@article{wang2025internvl35,
  title={Internvl3.5: Advancing open-source multimodal models in versatility, reasoning, and efficiency},
  author={Wang, Weiyun and Gao, Zhangwei and Gu, Lixin and Pu, Hengjun and Cui, Long and Wei, Xingguang and Liu, Zhaoyang and Jing, Linglin and Ye, Shenglong and Shao, Jie and others},
  journal={arXiv preprint arXiv:2508.18265},
  year={2025}
}

@misc{zhang2024llavanext,
  title={Llava-next: A strong zero-shot video understanding model},
  url={https://llava-vl.github.io/blog/2024-04-30-llava-next-video/},
  author={Zhang, Yuanhan and Li, Bo and Liu, haotian and Lee, Yong jae and Gui, Liangke and Fu, Di and Feng, Jiashi and Liu, Ziwei and Li, Chunyuan},
  month={April},
  year={2024}
}

@article{miao2025towards,
  title={Towards Physically Executable 3D Gaussian for Embodied Navigation},
  author={Miao, Bingchen and Wei, Rong and Ge, Zhiqi and Gao, Shiqi and Zhu, Jingzhe and Wang, Renhan and Tang, Siliang and Xiao, Jun and Tang, Rui and Li, Juncheng and others},
  journal={arXiv preprint arXiv:2510.21307},
  year={2025}
}

@inproceedings{yang2025thinking,
  title={Thinking in space: How multimodal large language models see, remember, and recall spaces},
  author={Yang, Jihan and Yang, Shusheng and Gupta, Anjali W and Han, Rilyn and Fei-Fei, Li and Xie, Saining},
  booktitle={Proceedings of the Computer Vision and Pattern Recognition Conference},
  pages={10632--10643},
  year={2025}
}

@article{xu2025spatialbench,
  title={Spatialbench: Benchmarking multimodal large language models for spatial cognition},
  author={Xu, Peiran and Wang, Sudong and Zhu, Yao and Li, Jianing and Qi, Gege and Zhang, Yunjian},
  journal={arXiv preprint arXiv:2511.21471},
  year={2025}
}

@inproceedings{wu2025difix3d+,
  title={Difix3d+: Improving 3d reconstructions with single-step diffusion models},
  author={Wu, Jay Zhangjie and Zhang, Yuxuan and Turki, Haithem and Ren, Xuanchi and Gao, Jun and Shou, Mike Zheng and Fidler, Sanja and Gojcic, Zan and Ling, Huan},
  booktitle={Proceedings of the IEEE/CVF Conference on Computer Vision and Pattern Recognition},
  pages={26024--26035},
  year={2025}
}

@inproceedings{held2025triangle,
  title={Triangle splatting for real-time radiance field rendering},
  author={Held, Jan and Vandeghen, Renaud and Deliege, Adrien and Hamdi, Abdullah and Rebain, Daniel and Giancola, Silvio and Cioppa, Anthony and Vedaldi, Andrea and Ghanem, Bernard and Tagliasacchi, Andrea and others},
  booktitle={Thirteenth International Conference on 3D Vision},
  year={2025}
}

@article{gao2024embodiedcity,
  title={Embodiedcity: A benchmark platform for embodied agent in real-world city environment},
  author={Gao, Chen and Zhao, Baining and Zhang, Weichen and Mao, Jinzhu and Zhang, Jun and Zheng, Zhiheng and Man, Fanhang and Fang, Jianjie and Zhou, Zile and Cui, Jinqiang and others},
  journal={arXiv preprint arXiv:2410.09604},
  year={2024}
}

@article{yang2025navspace,
  title={NavSpace: How Navigation Agents Follow Spatial Intelligence Instructions},
  author={Yang, Haolin and Long, Yuxing and Yu, Zhuoyuan and Yang, Zihan and Wang, Minghan and Xu, Jiapeng and Wang, Yihan and Yu, Ziyan and Cai, Wenzhe and Kang, Lei and others},
  journal={arXiv preprint arXiv:2510.08173},
  year={2025}
}

@article{gao2025octonav,
  title={Octonav: Towards generalist embodied navigation},
  author={Gao, Chen and Jin, Liankai and Peng, Xingyu and Zhang, Jiazhao and Deng, Yue and Li, Annan and Wang, He and Liu, Si},
  journal={arXiv preprint arXiv:2506.09839},
  year={2025}
}

@inproceedings{song2025towards,
  title={Towards long-horizon vision-language navigation: Platform, benchmark and method},
  author={Song, Xinshuai and Chen, Weixing and Liu, Yang and Chen, Weikai and Li, Guanbin and Lin, Liang},
  booktitle={Proceedings of the IEEE/CVF Conference on Computer Vision and Pattern Recognition},
  pages={12078--12088},
  year={2025}
}

@inproceedings{khanna2024goat,
  title={Goat-bench: A benchmark for multi-modal lifelong navigation},
  author={Khanna, Mukul and Ramrakhya, Ram and Chhablani, Gunjan and Yenamandra, Sriram and Gervet, Theophile and Chang, Matthew and Kira, Zsolt and Chaplot, Devendra Singh and Batra, Dhruv and Mottaghi, Roozbeh},
  booktitle={Proceedings of the IEEE/CVF Conference on Computer Vision and Pattern Recognition},
  pages={16373--16383},
  year={2024}
}

@inproceedings{yokoyama2024hm3d,
  title={Hm3d-ovon: A dataset and benchmark for open-vocabulary object goal navigation},
  author={Yokoyama, Naoki and Ramrakhya, Ram and Das, Abhishek and Batra, Dhruv and Ha, Sehoon},
  booktitle={2024 IEEE/RSJ International Conference on Intelligent Robots and Systems (IROS)},
  pages={5543--5550},
  year={2024},
  organization={IEEE}
}

@article{qiao2026navbench,
  title={Navbench: Probing multimodal large language models for embodied navigation},
  author={Qiao, Yanyuan and Hong, Haodong and Lyu, Wenqi and An, Dong and Zhang, Siqi and Xie, Yutong and Wang, Xinyu and Wu, Qi},
  journal={Advances in Neural Information Processing Systems},
  volume={38},
  pages={95365--95385},
  year={2026}
}

@inproceedings{qi2020reverie,
  title={Reverie: Remote embodied visual referring expression in real indoor environments},
  author={Qi, Yuankai and Wu, Qi and Anderson, Peter and Wang, Xin and Wang, William Yang and Shen, Chunhua and Hengel, Anton van den},
  booktitle={Proceedings of the IEEE/CVF conference on computer vision and pattern recognition},
  pages={9982--9991},
  year={2020}
}

@article{wu2025indoor,
  title={From indoor to open world: Revealing the spatial reasoning gap in mllms},
  author={Wu, Mingrui and Wang, Zhaozhi and Wang, Fangjinhua and Yang, Jiaolong and Pollefeys, Marc and Zhang, Tong},
  journal={arXiv preprint arXiv:2512.19683},
  year={2025}
}

@article{gholami2025spatial,
  title={Spatial reasoning with vision-language models in ego-centric multi-view scenes},
  author={Gholami, Mohsen and Rezaei, Ahmad and Weimin, Zhou and Mao, Sitong and Zhou, Shunbo and Zhang, Yong and Akbari, Mohammad},
  journal={arXiv preprint arXiv:2509.06266},
  year={2025}
}

@inproceedings{li2025sti,
  title={Sti-bench: Are mllms ready for precise spatial-temporal world understanding?},
  author={Li, Yun and Zhang, Yiming and Lin, Tao and Liu, XiangRui and Cai, Wenxiao and Liu, Zheng and Zhao, Bo},
  booktitle={Proceedings of the IEEE/CVF International Conference on Computer Vision},
  pages={5622--5632},
  year={2025}
}

@article{ren2025octree,
  title={Octree-GS: Towards Consistent Real-time Rendering with LOD-Structured 3D Gaussians},
  author={Ren, Kerui and Jiang, Lihan and Lu, Tao and Yu, Mulin and Xu, Linning and Ni, Zhangkai and Dai, Bo},
  journal={IEEE Transactions on Pattern Analysis and Machine Intelligence},
  year={2025},
  publisher={IEEE}
}

@inproceedings{chang2017matterport3d,
  title={Matterport3D: Learning from RGB-D Data in Indoor Environments},
  author={Chang, Angel and Dai, Angela and Funkhouser, Thomas and Halber, Maciej and Niebner, Matthias and Savva, Manolis and Song, Shuran and Zeng, Andy and Zhang, Yinda},
  booktitle={2017 International Conference on 3D Vision},
  pages={667--676},
  year={2017},
  organization={IEEE}
}

@inproceedings{chen2019touchdown,
  title={Touchdown: Natural language navigation and spatial reasoning in visual street environments},
  author={Chen, Howard and Suhr, Alane and Misra, Dipendra and Snavely, Noah and Artzi, Yoav},
  booktitle={Proceedings of the IEEE/CVF conference on computer vision and pattern recognition},
  pages={12538--12547},
  year={2019}
}

@article{kerbl20233dgs,
  title={3d gaussian splatting for real-time radiance field rendering},
  author={Kerbl, Bernhard and Kopanas, Georgios and Leimk{\"u}hler, Thomas and Drettakis, George and others},
  journal={ACM Trans. Graph.},
  volume={42},
  number={4},
  pages={139--1},
  year={2023}
}

@inproceedings{savva2019habitat,
  title={Habitat: A platform for embodied ai research},
  author={Savva, Manolis and Kadian, Abhishek and Maksymets, Oleksandr and Zhao, Yili and Wijmans, Erik and Jain, Bhavana and Straub, Julian and Liu, Jia and Koltun, Vladlen and Malik, Jitendra and others},
  booktitle={Proceedings of the IEEE/CVF international conference on computer vision},
  pages={9339--9347},
  year={2019}
}

@article{kolve2017ai2,
  title={Ai2-thor: An interactive 3d environment for visual ai},
  author={Kolve, Eric and Mottaghi, Roozbeh and Han, Winson and VanderBilt, Eli and Weihs, Luca and Herrasti, Alvaro and Deitke, Matt and Ehsani, Kiana and Gordon, Daniel and Zhu, Yuke and others},
  journal={arXiv preprint arXiv:1712.05474},
  year={2017}
}

@article{dongfang2025are,
  title={Are multimodal large language models ready for omnidirectional spatial reasoning?},
  author={Dongfang, Zihao and Zheng, Xu and Weng, Ziqiao and Lyu, Yuanhuiyi and Paudel, Danda Pani and Van Gool, Luc and Yang, Kailun and Hu, Xuming},
  journal={arXiv preprint arXiv:2505.11907},
  year={2025}
}

@inproceedings{johnson2017clevr,
  title={Clevr: A diagnostic dataset for compositional language and elementary visual reasoning},
  author={Johnson, Justin and Hariharan, Bharath and Van Der Maaten, Laurens and Fei-Fei, Li and Lawrence Zitnick, C and Girshick, Ross},
  booktitle={Proceedings of the IEEE conference on computer vision and pattern recognition},
  pages={2901--2910},
  year={2017}
}

@article{liu2025can,
  title={Can Multimodal Large Language Models Understand Spatial Relations?},
  author={Liu, Jingping and Liu, Ziyan and Cen, Zhedong and Zhou, Yan and Zou, Yinan and Zhang, Weiyan and Jiang, Haiyun and Ruan, Tong},
  journal={arXiv preprint arXiv:2505.19015},
  year={2025}
}

@inproceedings{ramakrishnan2025does,
  title={Does spatial cognition emerge in frontier models?},
  author={Ramakrishnan, Santhosh Kumar and Wijmans, Erik and Kraehenbuehl, Philipp and Koltun, Vladlen},
  booktitle={International Conference on Learning Representations},
  volume={2025},
  pages={101439--101498},
  year={2025}
}

@inproceedings{shiri2024empirical,
  title={An empirical analysis on spatial reasoning capabilities of large multimodal models},
  author={Shiri, Fatemeh and Guo, Xiao-Yu and Far, Mona Golestan and Yu, Xin and Haf, Reza and Li, Yuan-Fang},
  booktitle={Proceedings of the 2024 Conference on Empirical Methods in Natural Language Processing},
  pages={21440--21455},
  year={2024}
}

@article{stogiannidis2025mind,
  title={Mind the gap: Benchmarking spatial reasoning in vision-language models},
  author={Stogiannidis, Ilias and McDonagh, Steven and Tsaftaris, Sotirios A},
  journal={arXiv preprint arXiv:2503.19707},
  year={2025}
}

@article{tang2024sparkle,
  title={Sparkle: Mastering Basic Spatial Capabilities in Vision Language Models Elicits Generalization to Spatial Reasoning},
  author={Tang, Yihong and Qu, Ao and Wang, Zhaokai and Zhuang, Dingyi and Wu, Zhaofeng and Ma, Wei and Wang, Shenhao and Zheng, Yunhan and Zhao, Zhan and Zhao, Jinhua},
  journal={arXiv preprint arXiv:2410.16162},
  year={2024}
}

@inproceedings{wang2025spatial457,
  title={Spatial457: A diagnostic benchmark for 6d spatial reasoning of large mutimodal models},
  author={Wang, Xingrui and Ma, Wufei and Zhang, Tiezheng and de Melo, Celso M and Chen, Jieneng and Yuille, Alan},
  booktitle={Proceedings of the Computer Vision and Pattern Recognition Conference},
  pages={24669--24679},
  year={2025}
}

@article{yao2025lens,
  title={Lens: Multi-level evaluation of multimodal reasoning with large language models},
  author={Yao, Ruilin and Zhang, Bo and Huang, Jirui and Long, Xinwei and Zhang, Yifang and Zou, Tianyu and Wu, Yufei and Su, Shichao and Xu, Yifan and Zeng, Wenxi and others},
  journal={arXiv preprint arXiv:2505.15616},
  year={2025}
}

@article{yang2025mmsi,
  title={Mmsi-bench: A benchmark for multi-image spatial intelligence},
  author={Yang, Sihan and Xu, Runsen and Xie, Yiman and Yang, Sizhe and Li, Mo and Lin, Jingli and Zhu, Chenming and Chen, Xiaochen and Duan, Haodong and Yue, Xiangyu and others},
  journal={arXiv preprint arXiv:2505.23764},
  year={2025}
}

@inproceedings{yeh2026seeing,
  title={Seeing from another perspective: Evaluating multi-view understanding in mllms},
  author={Yeh, Chun-Hsiao and Wang, Chenyu and Tong, Shengbang and Cheng, Ta-Ying and Wang, Ruoyu and Chu, Tianzhe and Zhai, Yuexiang and Chen, Yubei and Gao, Shenghua and Ma, Yi},
  booktitle={Proceedings of the AAAI Conference on Artificial Intelligence},
  volume={40},
  number={14},
  pages={12000--12008},
  year={2026}
}

@article{gong2025space,
  title={Space-10: A comprehensive benchmark for multimodal large language models in compositional spatial intelligence},
  author={Gong, Ziyang and Li, Wenhao and Ma, Oliver and Li, Songyuan and Wang, Zhaokai and Li, Songyuan and Ji, Jiayi and Yang, Xue and Luo, Gen and Yan, Junchi and Ji, Rongrong},
  journal={arXiv preprint arXiv:2506.07966},
  year={2025}
}

@inproceedings{zhang2025do,
  title={Do vision-language models represent space and how? evaluating spatial frame of reference under ambiguities},
  author={Zhang, Zheyuan and Hu, Fengyuan and Lee, Jayjun and Shi, Freda and Kordjamshidi, Parisa and Chai, Joyce and Ma, Ziqiao},
  booktitle={The Thirteenth International Conference on Learning Representations (ICLR)},
  year={2025}
}

@inproceedings{zhou2025vlm4d,
  title={Vlm4d: Towards spatiotemporal awareness in vision language models},
  author={Zhou, Shijie and Vilesov, Alexander and He, Xuehai and Wan, Ziyu and Zhang, Shuwang and Nagachandra, Aditya and Chang, Di and Chen, Dongdong and Wang, Xin Eric and Kadambi, Achuta},
  booktitle={Proceedings of the IEEE/CVF international conference on computer vision},
  pages={8600--8612},
  year={2025}
}

@article{robobrain2,
  title={RoboBrain 2.0 Technical Report},
  author={{BAAI RoboBrain Team}},
  journal={arXiv preprint arXiv:2507.02029},
  year={2025}
}

@article{tan2026robobrain,
  title={RoboBrain 2.5: Depth in Sight, Time in Mind},
  author={Tan, Huajie and Zhou, Enshen and Li, Zhiyu and Xu, Yijie and Ji, Yuheng and Chen, Xiansheng and Chi, Cheng and Wang, Pengwei and Jia, Huizhu and Ao, Yulong and Cao, Mingyu and Chen, Sixiang and Li, Zhe and Liu, Mengzhen and Wang, Zixiao and Rong, Shanyu and Lyu, Yaoxu and Zhao, Zhongxia and Co, Peterson and Li, Yibo and Han, Yi and Xie, Shaoxuan and Yao, Guocai and Wang, Songjing and Zhang, Leiduo and Yang, Xi and Jiao, Yance and Shi, Donghai and Xie, Kunchang and Nie, Shaokai and Men, Chunlei and Lin, Yonghua and Wang, Zhongyuan and Huang, Tiejun and Zhang, Shanghang},
  journal={arXiv preprint arXiv:2601.14352},
  year={2026}
}

@misc{google2026geminiroboticser16,
  author       = {{Google DeepMind}},
  title        = {Gemini Robotics-ER 1.6},
  year         = {2026},
  month        = apr,
  howpublished = {Model Card},
  url          = {https://deepmind.google/models/model-cards/gemini-robotics-er-1-6/}
}

@article{oquab2023dinov2,
  title={Dinov2: Learning robust visual features without supervision},
  author={Oquab, Maxime and Darcet, Timoth{\'e}e and Moutakanni, Th{\'e}o and Vo, Huy and Szafraniec, Marc and Khalidov, Vasil and Fernandez, Pierre and Haziza, Daniel and Massa, Francisco and El-Nouby, Alaaeldin and others},
  journal={arXiv preprint arXiv:2304.07193},
  year={2023}
}

@inproceedings{kwon2023efficient,
  title={Efficient Memory Management for Large Language Model Serving with PagedAttention},
  author={Woosuk Kwon and Zhuohan Li and Siyuan Zhuang and Ying Sheng and Lianmin Zheng and Cody Hao Yu and Joseph E. Gonzalez and Hao Zhang and Ion Stoica},
  booktitle={Proceedings of the ACM SIGOPS 29th Symposium on Operating Systems Principles},
  year={2023}
}

@misc{google2025gemini3flash,
  author       = {{Google DeepMind}},
  title        = {Gemini 3 Flash Model Card},
  year         = {2025},
  month        = dec,
  howpublished = {Model card},
  url          = {https://storage.googleapis.com/deepmind-media/Model-Cards/Gemini-3-Flash-Model-Card.pdf}
}

@article{singh2025gpt5,
  title={Openai gpt-5 system card},
  author={Singh, Aaditya and Fry, Adam and Perelman, Adam and Tart, Adam and Ganesh, Adi and El-Kishky, Ahmed and McLaughlin, Aidan and Low, Aiden and Ostrow, AJ and Ananthram, Akhila and others},
  journal={arXiv preprint arXiv:2601.03267},
  year={2025}
}

@misc{qwen2026qwen35,
    title  = {{Qwen3.5}: Towards Native Multimodal Agents},
    author = {{Qwen Team}},
    month  = {February},
    year   = {2026},
    url    = {https://qwen.ai/blog?id=qwen3.5}
}

@article{bai2025qwen3vl,
  title={Qwen3-vl technical report},
  author={Bai, Shuai and Cai, Yuxuan and Chen, Ruizhe and Chen, Keqin and Chen, Xionghui and Cheng, Zesen and Deng, Lianghao and Ding, Wei and Gao, Chang and Ge, Chunjiang and others},
  journal={arXiv preprint arXiv:2511.21631},
  year={2025}
}

@article{team2026gemma4,
  title={Gemma 4 technical report},
  author={Team, Gemma and Abd, Sherif El and Aggarwal, Vaibhav and Algayres, Robin and Andreev, Alek and Bachem, Olivier and Ballantyne, Ian and Brick, Cormac and C{\u{a}}rbune, Victor and Casbon, Michelle and others},
  journal={arXiv preprint arXiv:2607.02770},
  year={2026}
}

@misc{internvla-n1,
    title = {{InternVLA-N1: An} Open Dual-System Navigation Foundation Model with Learned Latent Plans},
    author = {InternVLA-N1 Team},
    year = {2025},
    booktitle={arXiv},
}
}


\newpage

\appendix

\section{The \textit{Lost on Campus} Benchmark Construction Details}
\label{append:benchmark_construction}

\subsection{Environment Reconstruction}
\label{app:environment}

\paragraph{Data Sanitization and Privacy Compliance}
To ensure the visual fidelity of the reconstructed simulation and prevent the introduction of training artifacts, we perform a systematic filtering of the projected perspective images prior to reconstruction.  Specifically, the bottom-facing projections, which inevitably capture the static human operator during the data collection process, are entirely discarded. We follow a strict data anonymization protocol across the entire dataset. All collected visual frames are reviewed to identify and obfuscate Personally Identifiable Information (PII).  
All captures were collected in publicly accessible outdoor areas with approval, and faces and license plates are detected and obfuscated before reconstruction and release.

\paragraph{Environment Reconstruction}
Regarding the topological scope of the reconstructed scenes, our pipeline is designed to ensure geometric consistency across different evaluation scales. As introduced in Section~\ref{sec:environment}, \textit{Building-level environments} are generated by utilizing the comprehensive set of images captured along the entire perimeter of large architectural structures. These extensive reconstructions provide the facility to evaluate VLM agents on complex, long-horizon spatial instructions. To facilitate localized evaluations and create smaller-scale environments (\textit{e.g.}, specific intersections or localized pathways), rather than independently training separate 3DGS models for these smaller zones, we directly extract and isolate the relevant sub-regions from those larger-scale reconstructions.

\subsection{Navigation Case Construction}
\label{append:nav_case_construction}

 We automatically construct the benchmark from the training camera trajectories of each 3D Gaussian Splatting scene. The construction procedure is as follows.

 \textbf{Candidate pair generation.} We enumerate candidate start--goal view pairs from the training trajectory and retain only those whose 3D Euclidean distance lies between 5\,m and 25\,m. To avoid large elevation changes, we further require the vertical displacement between
  the two views, measured along the local up direction of the start view, to be no greater than 1\,m. 
  
  \textbf{Distance-stratified sampling.} From the valid candidate pool, we sample the final evaluation pairs using distance-stratified sampling. Specifically, we partition the 5--25\,m start--goal distance range into 20 equal-width intervals of 1\,m each, and select 100 start--goal pairs so that the number of sampled pairs is approximately balanced across intervals. This reduces bias toward any particular travel distance and ensures a more even coverage of the full 5--25\,m distance range. By
  default, each view is used at most once across all selected pairs, preventing repeated reuse of the same camera observation in different benchmark instances. After sampling, the direction of each selected pair is randomized with probability
  0.5, so either endpoint may serve as the starting view.
  
  \textbf{Pose perturbation and geometric constraints.} For each selected endpoint, we generate a start viewpoint by perturbing the original camera pose and re-rendering the scene from the perturbed pose. In our setting, the camera orientation is randomly rotated around its local up axis,
  with the rotation angle sampled uniformly from $[-180^\circ, 180^\circ]$. We do not apply additional horizontal translation perturbation. To further reduce height variation, we project each goal position onto the horizontal plane of the
  corresponding start position so that the start and goal share the same elevation. In addition, the goal camera is roughly oriented in the direction from the start to the goal (within $\pm90^\circ$), which excludes degenerate cases in which the goal view
  is opposite to the intended navigation direction.

\subsection{Spatial QA Construction}
\label{app:diagnostic_qa_construction}
\begin{table}[t]
\centering
\caption{
\textbf{Organization of the six diagnostic probes.}
The probes are organized along three aspects of the embodied perception--action loop.
}
\label{tab:qa_taxonomy}

\small
\setlength{\tabcolsep}{4.5pt}
\renewcommand{\arraystretch}{1.18}

\begin{tabularx}{\linewidth}{
    @{}
    p{0.26\linewidth}
    p{0.2\linewidth}
    X
    @{}
}
\toprule
\textbf{ESR Aspect}
& \textbf{Diagnostic Probe}
& \textbf{Detail} \\
\midrule

\multirow[t]{2}{=}{Coupling actions to observations}
& \textbf{Action Grounding}
& Relating visual transitions to the agent's own actions \\
\cmidrule(lr){2-3}

& \textbf{Spatial Foresight}
& Predicting future visual changes induced by actions \\

\midrule

\multirow[t]{3}{=}{Maintaining spatial state during movement}
& \textbf{Metric Awareness}
& Estimating relative distance and spatial progress \\
\cmidrule(lr){2-3}

& \textbf{Ego-allocentric \newline Localization}
& Anchoring egocentric observations in a global spatial frame \\
\cmidrule(lr){2-3}

& \textbf{Spatio-temporal \newline Consistency}
& Maintaining a consistent representation of spatial relations under ego-motion \\

\midrule

Relating observations and spatial states to goals
& \textbf{Goal-directed \newline Planning}
& Inferring implicit route structure from observations and reasoning about spatial relations among current, intermediate, and goal viewpoints \\

\bottomrule
\end{tabularx}
\end{table}

As shown in Table~\ref{tab:qa_taxonomy}, we organize the six diagnostic tasks along the embodied perception--action loop. 
Closed-loop embodied spatial reasoning requires the agent to (i) couple actions to observations, 
(ii) maintain spatial state during movement, and 
(iii) relate observations and spatial states to goals. 
Accordingly, the six tasks are designed as complementary diagnostic probes spanning these aspects and different levels of analysis, rather than as an exhaustive or orthogonal decomposition of ESR.

We automatically construct the diagnostic spatial QA protocol from navigation rollouts produced by different VLM agents. Following Section~\ref{sec:diagnostic_qa}, the protocol decomposes embodied spatial reasoning into six diagnostic tasks, each targeting a distinct spatial capability. For each task, we first collect candidate samples from rollout trajectories under task-specific temporal and geometric constraints, then generate the correct answer together with plausible distractors, and finally subsample the resulting pool to balance scenes, source agents, and rollout cases.

\paragraph{Action Grounding.}
We parse each rollout into contiguous segments of valid primitive actions and aligned camera poses, and extract short windows of consecutive observations. A sample with $N$ observation frames is paired with the corresponding $N-1$ ground-truth primitive actions between adjacent frames. We only retain windows in which every transition between adjacent observations is associated with a valid primitive action and a temporally continuous, renderable pose sequence.

\paragraph{Spatial Foresight.}
We sample a start observation and a short action sequence from a rollout trajectory, and use the true post-action observation reached after executing the sequence as the correct answer. Each sample is converted into a four-way multiple-choice question by adding three distractors. The distractors are generated from three sources: an action sequence obtained by modifying a single action in the ground-truth sequence, a real pose from another step of the same rollout, and a random action sequence of the same length. All distractors are filtered to remain sufficiently separated from the ground-truth future pose and from one another.

\paragraph{Metric Awareness.}
We construct a scene-level pose pool and sample a goal pose as the distance reference. Each candidate observation is associated with its camera-center distance to the goal. We then generate two forms of questions. In trend questions, the sampled distances must satisfy one of four strict patterns: approaching, receding, approaching-then-receding, or receding-then-approaching. In extremum-selection questions, the model is asked to identify the closest or farthest frame in the sequence. Candidate sequences are further filtered by distance-gap, spatial-continuity, and view-consistency constraints.

\paragraph{Goal-directed Planning.}
We operate on the reconstructed camera trajectory of each scene and enumerate initial-goal pairs under distance and temporal-gap constraints. For each valid pair, the correct subgoal is sampled from the middle portion of the trajectory interval between the initial and goal viewpoints. Distractors are sampled from outside this interval while remaining spatially close to the endpoints. Additional directional constraints are imposed so that distractors remain visually plausible without lying on the true intermediate route.

\paragraph{Ego-allocentric Localization.}
We combine rollout poses with the original scene trajectory to generate BEV-based localization questions. We first identify valid initial-goal pairs from rollout trajectories under distance and directional consistency constraints, and then instantiate two task variants. In \emph{initial localization}, the correct option is the true initial position, while the distractors are drawn from one rollout pose and two scene-trajectory poses. In \emph{goal localization}, the correct option is the true goal position, while the distractors are drawn from one rollout pose, one scene-trajectory pose, and one local perturbation around the goal. All candidate markers are required to remain spatially separated on the map.

\paragraph{Spatio-temporal Consistency.}
We extract short consecutive observation windows from rollout trajectories. For each window, we render the first frame and detect a salient object, represented by a semantic tag and a 2D bounding box. Using the first-frame depth map, we estimate the object's 3D world position and project it through the full motion sequence. This process yields three labels: whether the object remains visible in the final frame, its relative direction with respect to the final camera, and its apparent horizontal motion pattern across the sequence. The final relative direction is discretized into eight camera-centric sectors. Depending on whether the horizontal motion signal is sufficiently clear, each sample is instantiated as either a visibility-and-motion question or a visibility-and-position question.

\begin{figure*}[ht]
    \centering

    \begin{subfigure}[t]{0.9\textwidth}
        \centering
        \includegraphics[width=\linewidth]{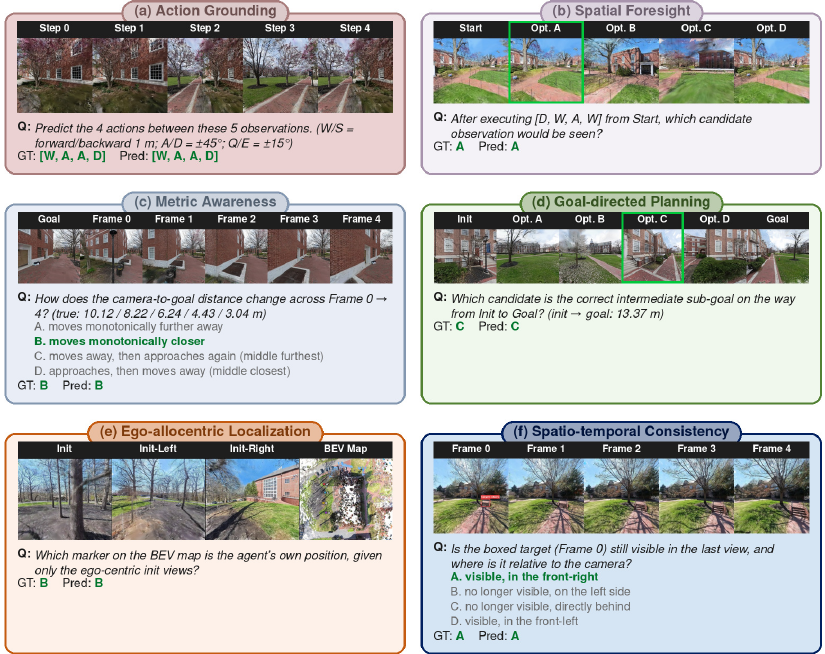}
        \caption{Representative diagnostic QA cases correctly answered by Gemini-3-Flash.}
        \label{fig:qa_correct}
    \end{subfigure}

    \vspace{0.15cm}
    
    \begin{subfigure}[t]{0.9\textwidth}
        \centering
        \includegraphics[width=\linewidth]{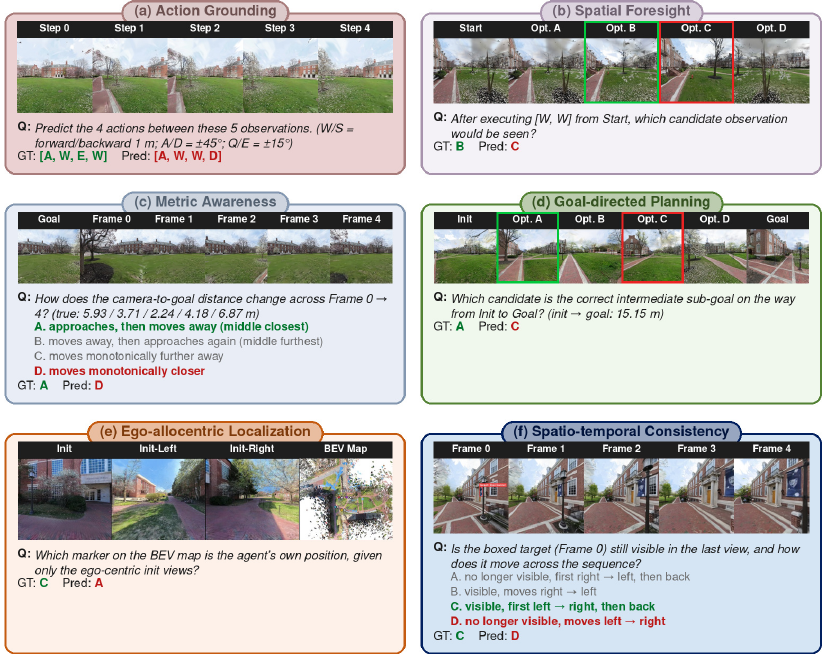}
        \caption{Representative diagnostic QA cases incorrectly answered by Gemini-3-Flash.}
        \label{fig:qa_incorrect}
    \end{subfigure}

    \caption{
    Representative diagnostic QA cases answered by Gemini-3-Flash. The examples correspond to Action Grounding, Spatial Foresight, Metric Awareness, Goal-directed Planning, Ego-allocentric Localization, and Spatio-temporal Consistency.
    }
    \label{fig:qa_cases}
\end{figure*}

\subsection{Qualitative Examples of Diagnostic Spatial QA}
\label{append:qa_examples}

Figure~\ref{fig:qa_cases} presents qualitative examples from the six diagnostic spatial QA tasks evaluated with Gemini-3-Flash. For each diagnostic probe, we show one correctly answered example and one incorrectly answered example. These cases illustrate both the task format and the types of spatial reasoning successes and failures captured by our diagnostic protocol.

\section{Reconstruction Analysis}
\label{append:reconstruction_analysis}
\begin{table}[t]
\centering
\caption{
\textbf{Reconstruction fidelity across the 15 reconstructed scenes.}
We report PSNR, SSIM, and LPIPS averaged over all views, the top 10\% views, and the best views.
}
\label{tab:reconstruction_fidelity}
\small
\resizebox{0.7\linewidth}{!}{
\begin{tabular}{lccc}
\toprule
\textbf{Views}
& \textbf{PSNR $\uparrow$}
& \textbf{SSIM $\uparrow$}
& \textbf{LPIPS $\downarrow$} \\
\midrule
Mean over all views
& 22.30 & 0.777 & 0.216 \\

Mean over top 10\% views
& 25.51 & 0.863 & 0.129 \\

Mean over best views
& 27.30 & 0.885 & 0.102 \\
\bottomrule
\end{tabular}
}
\end{table}

\begin{table}[t]
\centering
\caption{
\textbf{Artifact mitigation analysis with Difix3D+.}
Qwen3.5-27B is evaluated under the same settings on a scene-balanced subset of 150 navigation episodes. We compare the original 3DGS observations with observations enhanced by Difix3D+.
}
\label{tab:difix_analysis}
\small
\resizebox{0.75\linewidth}{!}{
\begin{tabular}{llcc}
\toprule
\textbf{Evaluation}
& \textbf{Metric}
& \textbf{Original 3DGS}
& \textbf{+ Difix3D+} \\
\midrule
Diagnostic QA
& Accuracy
& 59.0
& \textbf{61.0} \\

Closed-loop navigation
& SR@2.5
& 13.3
& \textbf{14.0} \\

Closed-loop navigation
& SR@5
& \textbf{36.7}
& 36.0 \\

\bottomrule
\end{tabular}
}
\end{table}

\begin{figure}[ht]
  \centering
  \includegraphics[width=\textwidth]{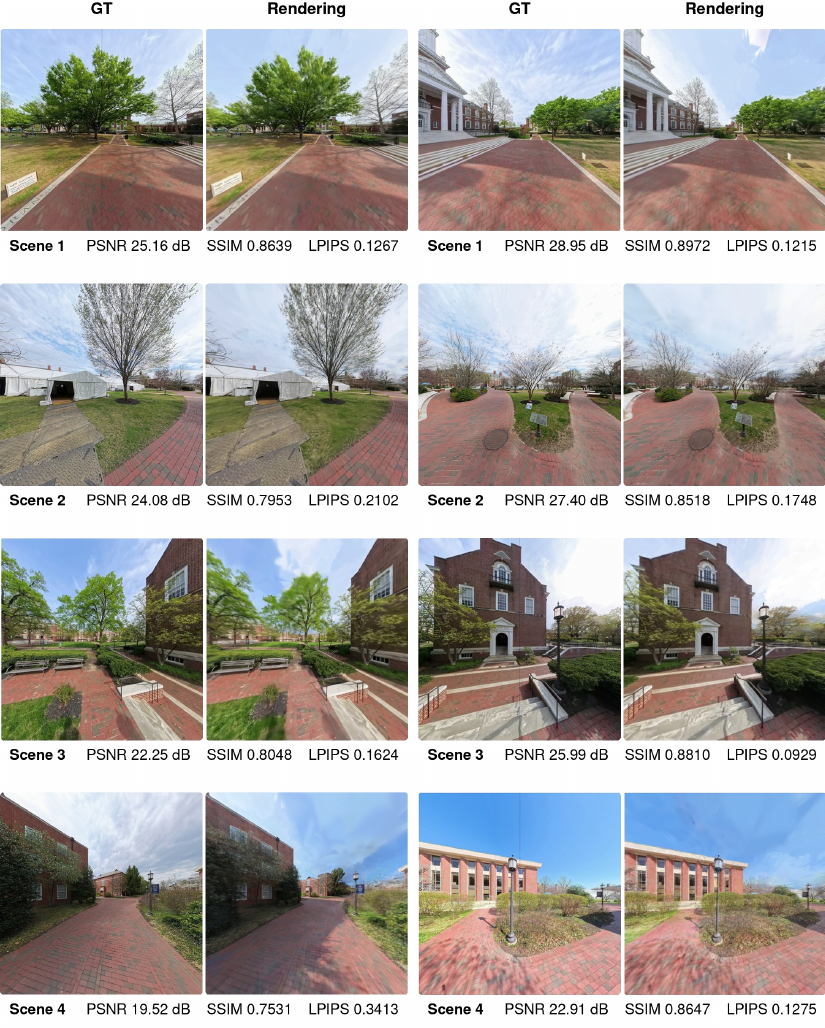}
  \caption{\textbf{Qualitative reconstruction fidelity.} For each of four representative scenes, we compare a real captured photograph (\textit{GT}) with the 3DGS rendering from the same camera pose (\textit{Rendering}), together with the per-view PSNR/SSIM/LPIPS. The four scenes span the range of reconstruction quality in our benchmark (scene-average PSNR $24.89$, $23.94$, $22.51$, and $18.71$\,dB). Visible discrepancies are concentrated in sky, lawn, and foliage textures, whereas building facades, road boundaries, walkways, and other navigation-relevant structures remain visually and geometrically consistent.}
  \label{fig:reconstruction_fidelity}
\end{figure}

\textbf{Reconstruction fidelity.}
To investigate whether reconstruction artifacts affect benchmark evaluation, we first evaluate the rendering quality over all 15 reconstructed scenes. As shown in Table~\ref{tab:reconstruction_fidelity}, the reconstructed environments achieve an average PSNR of 22.30 dB, SSIM of 0.777, and LPIPS of 0.216 over all views. We additionally report the average results over the top 10\% views and the best views to characterize the upper range of reconstruction quality. RTK/GPS calibration further anchors the reconstructed geometry and metric scale to the physical world. The qualitative visualization is shown in Figure~\ref{fig:reconstruction_fidelity}.

\textbf{Artifact mitigation with Difix3D+.}
To further validate the effect of reconstruction artifacts, we apply the inference-time enhancer from Difix3D+~\cite{wu2025difix3d+} to the 3DGS-rendered observations used in both diagnostic QA and closed-loop navigation. As shown in Table~\ref{tab:difix_analysis}, applying Difix3D+ yields only marginal changes: diagnostic QA accuracy improves from 59.0 to 61.0, SR@2.5 from 13.3 to 14.0, while SR@5 changes from 36.7 to 36.0. These results suggest that appearance-level reconstruction artifacts are not the primary driver of benchmark performance.

\begin{figure}[ht]
  \centering
  \includegraphics[width=\textwidth]{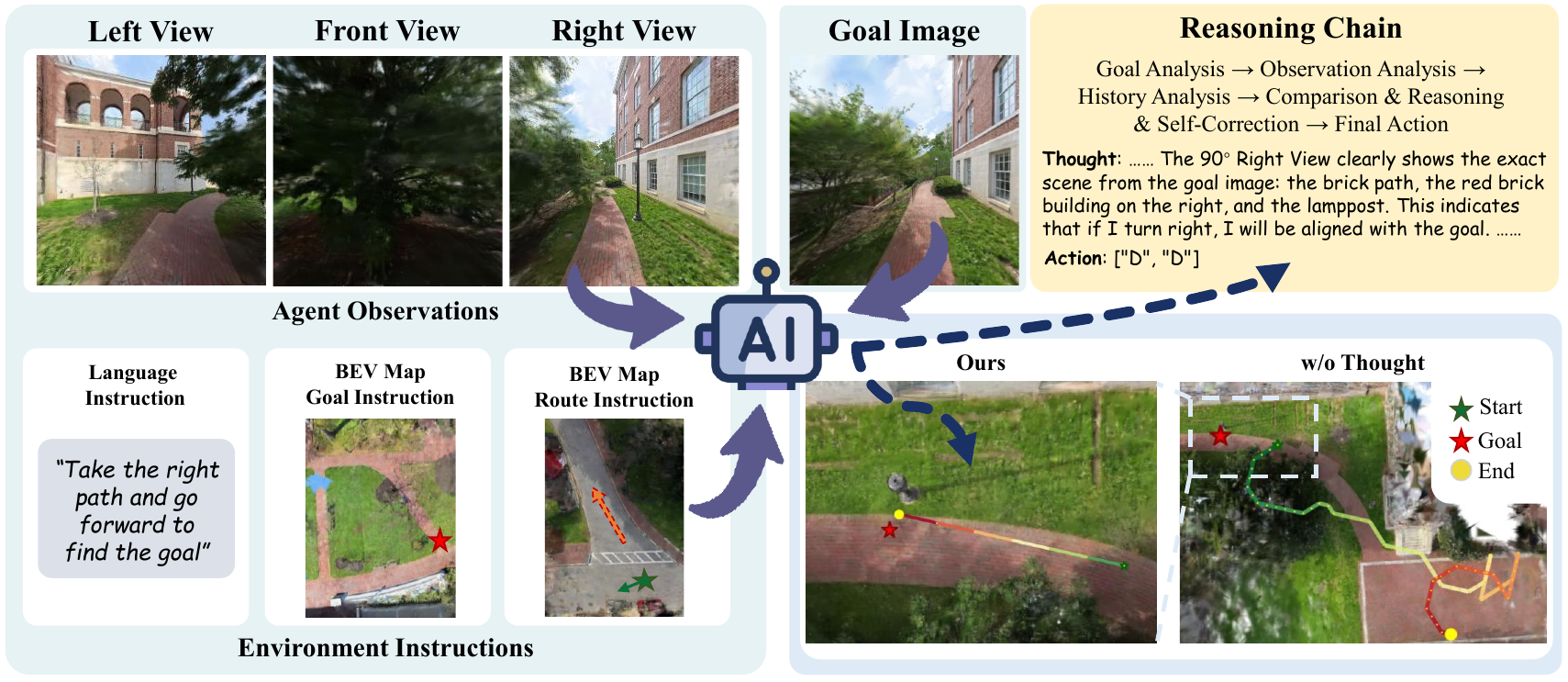}
  \caption{An illustration of evaluation input and case visualization. `BEV Map Goal Instruction' denotes Goal-Annotated BEV and `BEV Map Route Instruction' denotes Visual Route Map. }
  \label{fig:agent_illustration}
\end{figure}

\begin{figure}[ht]
  \centering
  \includegraphics[width=1.\textwidth]{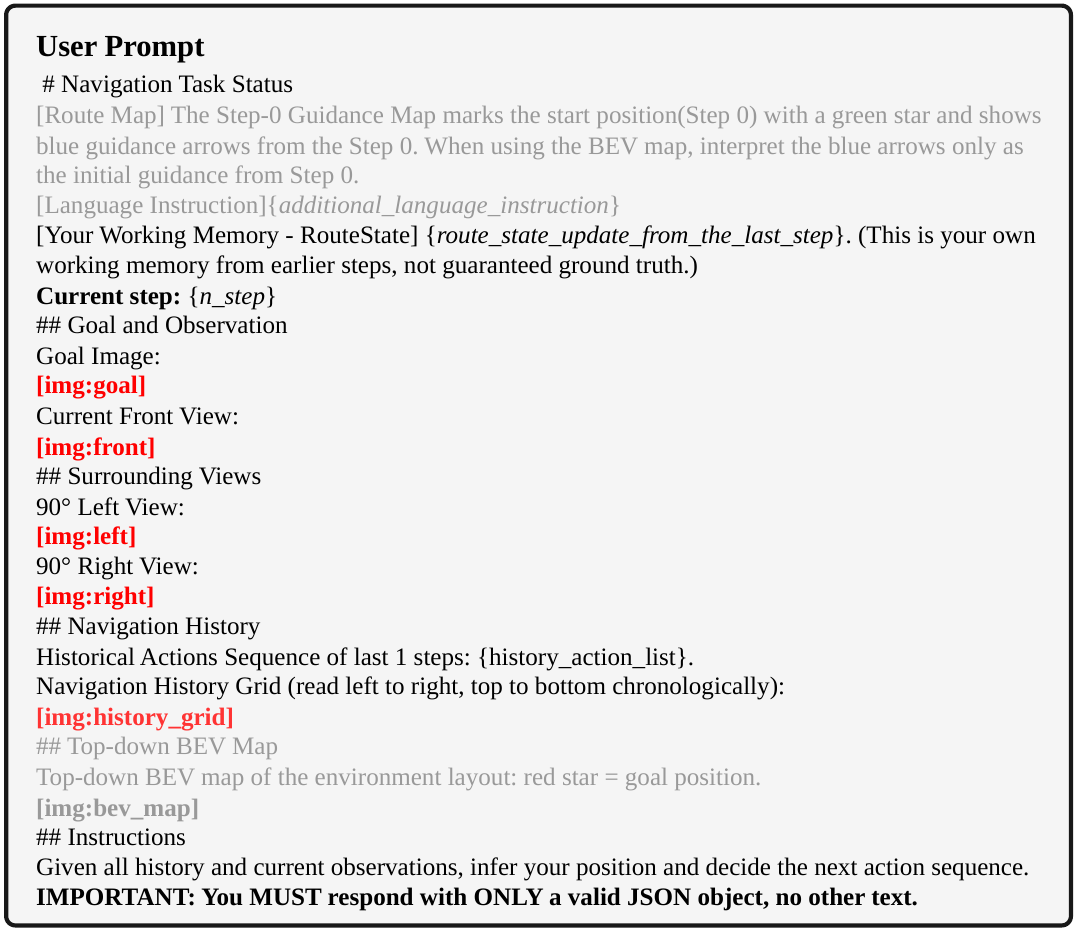}
  \caption{Unified user prompt template for navigation evaluation.  Gray text denotes optional setting-specific fields that are conditionally included according to the setting: 
\texttt{[Language Instruction]} block is used only in \textit{Language Instruction}; 
\texttt{Top-down BEV Map} block is used in both \textit{Goal-Annotated BEV} and \textit{Visual Route Map}; 
and \texttt{[Route Map]} is used only in \textit{Visual Route Map}.
}
  \label{fig:User_prompt}
\end{figure}

\section{Evaluation Setup}
\label{append:evaluation_setup}

\subsection{Implementation Details}
\label{append:implementation_details}

All open-source VLMs are served using the vLLM~\cite{kwon2023efficient} backend.
Experiments for open-source models are conducted on a single NVIDIA H100 GPU with 80GB VRAM.
To ensure reproducibility, unless otherwise specified, we adopt greedy decoding for all models, with temperature set to 0 and both top-$p$ and top-$k$ set to 1.
At each navigation step, all visual and textual inputs are organized into a structured prompt, as detailed in Figure~\ref{fig:agent_illustration}.

\subsection{Prompt Template for Navigation Task}

\begin{figure}[ht]
  \centering
  \includegraphics[width=1.0\textwidth]{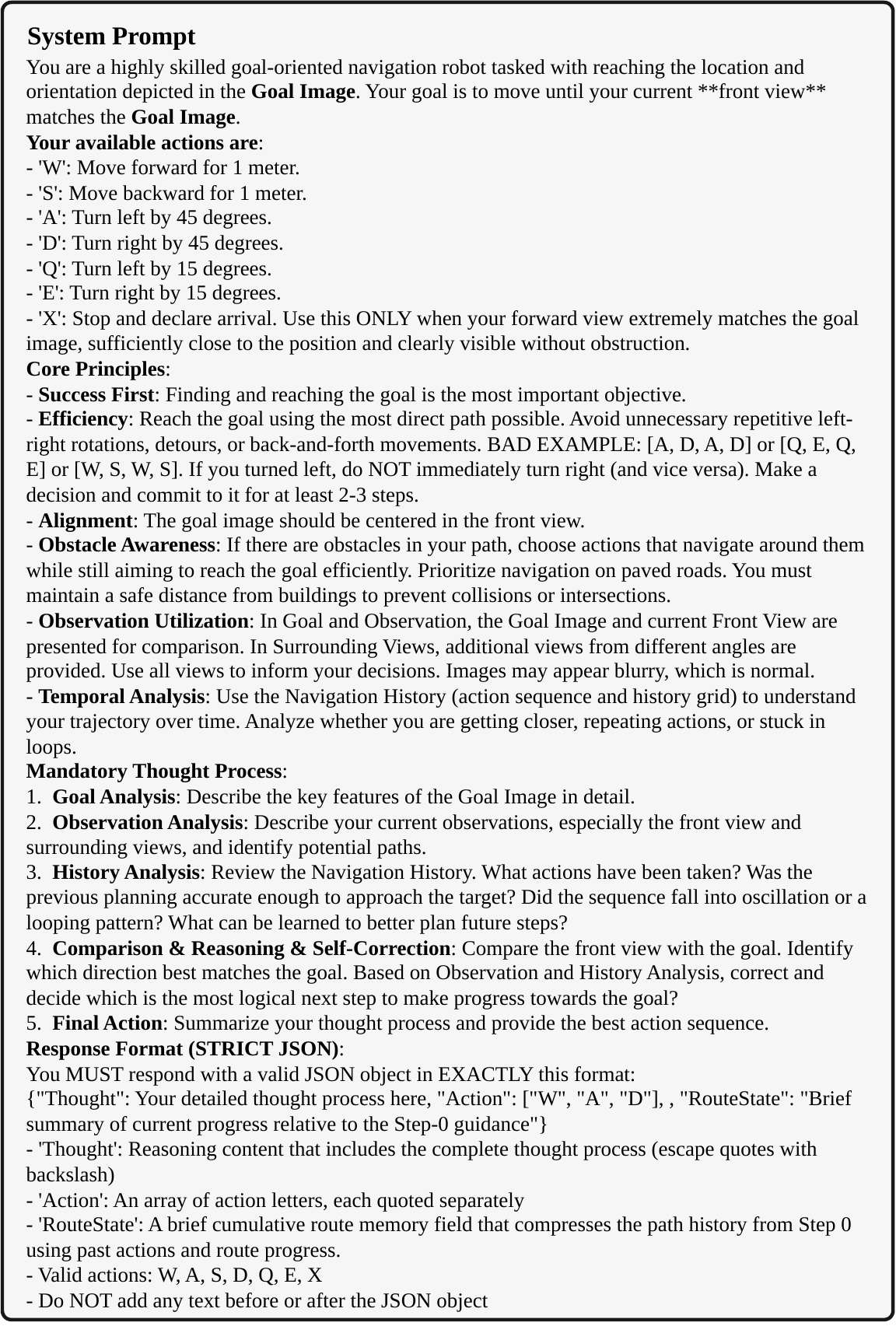}
  \caption{Unified system prompt template for navigation evaluation.
}
  \label{fig:sys_prompt}
\end{figure}

At each navigation step, we use structured prompting to obtain the model's action decision.
The system prompt specifies the required output format, which includes a \textit{thought} field and an \textit{action} field.
The task prompt provides the step-specific input to the agent, including the navigation history and the current visual observations.
The current observations consist of front, left, and right egocentric views, together with the goal image.
Depending on the evaluation setting, the prompt may additionally include a language instruction, a goal-annotated BEV image, or a visual route map.

Following prior works~\cite{gao2025octonav}, we retain a sliding window of historical observations and actions at each step.
The history window size is set to $N=8$. Figure~\ref{fig:User_prompt} and Figure~\ref{fig:sys_prompt} illustrate the complete prompt template for navigation evaluation.

\section{Additional Experimental Results}
\label{append:full_results}

\subsection{Analysis of Potential Shortcuts}
\label{append:shortcut_analysis}
\begin{table}[t]
\centering
\caption{
\textbf{Potential shortcut analysis for diagnostic QA.}
We compare Qwen3.5-27B under a text-only setting against majority-class and random-chance baselines, as well as the standard full-input setting.
}
\label{tab:shortcut_qa}
\resizebox{0.88\linewidth}{!}{
\begin{tabular}{lcccc}
\toprule
\textbf{Task}
& \textbf{Text-only}
& \textbf{Majority}
& \textbf{Chance}
& \textbf{Full Input} \\
\midrule
Action Grounding
& 44.2 & 46.6 & 16.7 & \textbf{82.5} \\

Spatial Foresight
& 26.5 & 27.2 & 25.0 & \textbf{55.3} \\

Metric Awareness
& 26.9 & 26.9 & 23.3 & \textbf{61.7} \\

Goal-directed Planning
& 25.7 & 26.9 & 25.0 & \textbf{72.2} \\

Ego-allocentric Localization
& 25.8 & 26.1 & 25.0 & \textbf{43.2} \\

Spatio-temporal Consistency
& 33.0 & 26.0 & 25.0 & \textbf{50.0} \\

\bottomrule
\end{tabular}
}
\end{table}

\begin{table}[t]
\centering
\caption{
\textbf{Potential shortcut analysis for navigation.}
All methods are evaluated on the same scene-balanced subset of 150 navigation episodes.
}
\label{tab:shortcut_navigation}
\resizebox{0.72\linewidth}{!}{
\begin{tabular}{lcccc}
\toprule
\textbf{Setting}
& \textbf{SR@2.5}
& \textbf{SR@5}
& \textbf{OSR@5}
& \textbf{DTG} \\
\midrule
Random policy
& 0.0 & 0.7 & 1.3 & 14.88 \\

Forward-biased random
& 0.0 & 2.7 & 5.3 & 15.82 \\

DINOv2 matching
& 3.3 & 6.7 & 6.7 & 13.97 \\

Qwen3.5-27B
& 13.3 & 36.7 & 39.3 & 9.44 \\

Oracle shortest-path
& \textbf{100.0}
& \textbf{100.0}
& \textbf{100.0}
& \textbf{1.03} \\

\bottomrule
\end{tabular}
}
\end{table}

\textbf{Diagnostic QA.}
To examine whether the diagnostic QA tasks can be solved through linguistic priors or label-distribution biases, we evaluate Qwen3.5-27B in a text-only setting where all visual inputs are removed. We additionally report a majority-class baseline that always predicts the most frequent answer label for each task, together with the corresponding random-chance accuracy. 
As shown in Table~\ref{tab:shortcut_qa}, text-only performance is largely comparable to that of Chance (random selection) or Majority class. Visual input provides a substantial improvement, which indicates that the QA cannot be solved based solely on linguistic priors or shortcuts.

\textbf{Navigation.} We further test a variety of non-VLM baselines, such as Random policy (randomly selecting one action), Forward-biased random (50\% probability of moving forward and the rest randomly), DINOv2 matching (selecting the action with the highest similarity), and Oracle shortest-path (calculating actions directly based on the true goal position).
As shown in Table~\ref{tab:shortcut_navigation}, this benchmark task cannot be solved through visual matching alone.

\subsection{Statistical Analysis}
\label{append:statistical_analysis}
\begin{table}[t]
\centering
\caption{
\textbf{Statistical analysis of navigation performance for Qwen3.5-27B.}
We report SR@2.5 and SR@5 with 95\% bootstrap confidence intervals on the full benchmark and across different scene types and start--goal distance ranges.
}
\label{tab:statistical_analysis}
\small
\resizebox{0.8\linewidth}{!}{
\begin{tabular}{lccc}
\toprule
\textbf{Subset}
& \textbf{Cases}
& \textbf{SR@2.5}
& \textbf{SR@5} \\
\midrule

Full benchmark
& 1500
& 9.93 [8.47, 11.47]
& 34.07 [31.67, 36.47] \\

\midrule
Building-level
& 1100
& 9.36 [7.64, 11.09]
& 32.91 [30.18, 35.73] \\

Intersection-level
& 400
& 11.50 [8.50, 14.75]
& 37.25 [32.50, 42.00] \\

\midrule
5--10 m
& 375
& 18.13 [14.40, 22.13]
& 59.47 [54.40, 64.53] \\

10--15 m
& 375
& 11.20 [8.00, 14.40]
& 37.07 [32.27, 41.87] \\

15--20 m
& 376
& 6.38 [3.99, 9.04]
& 22.34 [18.34, 26.60] \\

20--25 m
& 374
& 4.01 [2.14, 6.15]
& 17.38 [13.64, 21.39] \\

\bottomrule
\end{tabular}
}
\end{table}

To quantify statistical uncertainty, we compute 95\% bootstrap confidence intervals by repeated sampling with replacement for Qwen3.5-27B on the full 1,500-case navigation benchmark. The resulting intervals are 9.93 [8.47, 11.47] for SR@2.5, 34.07 [31.67, 36.47] for SR@5.
We further stratify navigation performance by scene type and start--goal distance. As shown in Table~\ref{tab:statistical_analysis}, performance is relatively similar across scene types, while success rates decrease consistently as the start--goal distance increases.

\subsection{Scale-Normalized Indoor Comparison}
\label{append:indoor_comparison}
\begin{table}[t]
\centering
\caption{
\textbf{Scale-normalized indoor image-goal navigation.} 150 Habitat episodes with 1--5\,m paths, 0.2\,m steps, and 0.5/1\,m thresholds. \textit{Strict} and \textit{Loose} denote the 0.5\,m and 1\,m thresholds, which are matched in units of action steps.
}
\label{tab:indoor_compare}
\vspace{0.1cm}

\setlength{\tabcolsep}{6pt}
\renewcommand{\arraystretch}{1.15}
\begin{tabular}{@{}ll cc cc cc@{}}
\toprule
\multirow{2}{*}{\textbf{Model}} 
  & \multicolumn{2}{c}{\textbf{SR}$\uparrow$}
  & \multicolumn{2}{c}{\textbf{OSR}$\uparrow$}
  & \multicolumn{2}{c}{\textbf{SPL}$\uparrow$} \\
\cmidrule(lr){2-3}\cmidrule(lr){4-5}\cmidrule(lr){6-7}
 & Strict & Loose & Strict & Loose & Strict & Loose \\
\midrule
Qwen3-VL-8B-Instruct  & 44.0  & 58.0  & 44.7  & 60.0  & 27.7 & 40.2  \\

Qwen3-VL-32B-Instruct & 63.3  & 70.0  & 64.7  & 70.7  & 39.5 & 44.0  \\

Qwen3.5-9B  & 28.0  & 41.3  & 28.7  & 42.0  & 19.4 & 30.6  \\

Qwen3.5-27B & 59.3  & 66.0  & 62.0  & 67.3  & 36.2 & 40.6  \\

\bottomrule
\end{tabular}
\end{table}

To examine whether the low outdoor success rates in Table~\ref{tab:nav_main} stem from the outdoor environments rather than from our evaluation protocol, we evaluate four open-source VLMs on a scale-normalized indoor counterpart in Habitat~\cite{savva2019habitat}. We use 150 image-goal episodes uniformly sampled from 15 validation scenes, with start--goal distances of 1--5\,m, a translation step of 0.2\,m, and success thresholds of 0.5\,m and 1\,m. These are one-fifth of the outdoor 5--25\,m paths, 1\,m step, and 2.5/5\,m thresholds, so both settings cover paths of 5--25 translation steps and apply the same thresholds when measured in steps. All other settings, including observations, rotation actions, the episode limit, prompts, and decoding, are identical to the outdoor evaluation.

As shown in Table~\ref{tab:indoor_compare}, indoor performance is higher than outdoor performance for every model on all six metrics. At the strict threshold, indoor SR exceeds outdoor SR by 21.7--57.0 points (e.g., 59.3 vs.~9.93 for Qwen3.5-27B), and at the loose threshold by 20.8--41.9 points. This comparison matches path lengths and success thresholds in units of translation steps, but not scene content or rendering. It therefore quantifies the overall difficulty gap between the two environment types rather than attributing it to a single factor such as semantic density.

\subsection{Comparison with Specialized Embodied and Navigation Models}
\label{append:specialized_baselines}
\begin{table}[t] 
\centering 
\caption{\textbf{Comparison with specialized embodied and navigation models on a fixed scene-balanced 150-episode subset.} The reference foundation models are re-evaluated on the same subset under identical observations, actions, and metrics.}
\label{tab:specialized_nav} 
\renewcommand{\arraystretch}{1.12}
\begin{tabular}{lccc} 
\toprule 
Model & SR@2.5 $\uparrow$ & SR@5 $\uparrow$ & OSR@5 $\uparrow$ \\ 
\midrule 
RoboBrain 2.0-7B & 1.33 & 5.33 & 6.67 \\ 
RoboBrain 2.0-32B & 2.67 & 10.00 & 48.67 \\ 
RoboBrain 2.5-8B-NV & 6.00 & 21.33 & 29.33 \\ 
InternVLA-N1 S1 & 8.00 & 12.00 & 18.67 \\ 
Qwen3.5-27B (reference) & 13.33 & 36.67 & 39.33 \\ 
Gemini Robotics-ER 1.6 Preview & 25.33 & 53.33 & 62.67 \\ 
Gemini-3-Flash (reference) & 26.00 & 51.33 & 66.00 \\ 
\bottomrule 
\end{tabular} 
\end{table}

We additionally evaluate embodied/spatial reasoning models under the same navigation protocol, including RoboBrain2.0-7B/32B~\cite{robobrain2},
RoboBrain2.5-8B-NV~\cite{tan2026robobrain},
InternVLA-N1~\cite{internvla-n1},
and Gemini Robotics-ER 1.6 Preview~\cite{google2026geminiroboticser16}. For a controlled comparison, all additional models and the representative general-purpose VLM are evaluated on the same uniformly sampled, scene-balanced 150-episode subset covering all 15 scenes. As shown in Table~\ref{tab:specialized_nav}, the results indicate that performance differences cannot be explained solely by model scale or embodied design: the RoboBrain series models all perform poorly, and Gemini Robotics-ER showed no clear advantage compared to the general-purpose Gemini-3-Flash. These results suggest that the difficulty of \textit{Lost on Campus} is not specific to prompting general-purpose VLMs, and that current specialized embodied models also face substantial challenges in continuous outdoor spatial reasoning.

\subsection{Full Controlled Setting Results}
\label{append:full_controlled_results}

\begin{table*}[t]
\centering
\small
\caption{\textbf{Additional outcome-oriented metrics for controlled spatial instruction variants.}
SPL is reported under 5.0m.
For each model, the best setting for each metric is marked in \textbf{bold}.}
\label{tab:controlled_outcome_metrics}
\setlength{\tabcolsep}{5.2pt}
\renewcommand{\arraystretch}{1.18}
\begin{tabular}{@{}l cc cc cc cc@{}}
\toprule
\multirow{2}{*}{\textbf{Setting}}
  & \multicolumn{2}{c}{\textbf{Qwen3-VL-8B}}
  & \multicolumn{2}{c}{\textbf{Qwen3.5-9B}}
  & \multicolumn{2}{c}{\textbf{Qwen3.5-27B}}
  & \multicolumn{2}{c}{\textbf{Qwen3-VL-32B}} \\
\cmidrule(lr){2-3}\cmidrule(lr){4-5}\cmidrule(lr){6-7}\cmidrule(lr){8-9}
  & SPL$\uparrow$ & DTG$\downarrow$
  & SPL$\uparrow$ & DTG$\downarrow$
  & SPL$\uparrow$ & DTG$\downarrow$
  & SPL$\uparrow$ & DTG$\downarrow$ \\
\midrule
\rowcolor{gray!12}
\textit{Image-Goal Nav.}
  & 3.50 & 27.18
  & 0.39 & 26.82
  & 3.59 & 23.55
  & 4.30 & 22.04 \\
\addlinespace[2pt]
\textit{w/ Language Instruction}
  & 2.82 & \textbf{25.37}
  & \textbf{2.48} & \textbf{25.91}
  & 4.67 & \textbf{21.98}
  & 5.35 & 21.50 \\
\textit{w/ Goal-Annotated BEV}
  & \textbf{4.13} & 25.62
  & 0.69 & 27.54
  & 6.09 & 22.06
  & \textbf{6.75} & \textbf{20.77} \\
\textit{w/ Visual Route Map}
  & 1.28 & 26.81
  & 0.98 & 26.83
  & \textbf{7.70} & 22.74
  & 5.25 & 23.53 \\
\bottomrule
\end{tabular}
\end{table*}

In the main text, we report \textit{SR} and \textit{OSR} as changes relative to the image-goal navigation setting.
In Table~\ref{tab:controlled_outcome_metrics}, we provide additional absolute metrics for the controlled spatial instruction variants, including \textit{SPL} under 5m threshold and \textit{DTG}. The configuration is the same as in Table~\ref{tab:nav_instruction}.

\textbf{VLMs struggle to convert linguistic spatial relations into interactive action execution.} Adding textual spatial instruction, which we hope to probe the agent's ability to convert egocentric linguistic spatial relations into embodied interaction, leads to limited meaningful gains. Among the models that benefit from this setting, the SR improvement ranges from $+0.78$ to $+2.35$, as shown in Table~\ref{tab:nav_instruction}. This suggests that VLMs may possess partial static understanding of language-based spatial relations, but such understanding does not reliably transfer to embodied action execution.

\textbf{Larger VLMs show emerging self-localization and active planning ability.}
The \textit{Goal-Annotated BEV} variant measures whether models can use global goal information to support self-localization and active planning during interaction. 
Compared with \textit{Image-Goal Nav.}, the \textit{Goal-Annotated BEV} setting yields the greatest SR improvements for the two larger models. Qwen3.5-27B improves from 3.92\% to 8.63\%, and Qwen3-VL-32B-Instruct improves from 5.88\% to 9.02\%.
It also produces the largest OSR gain for Qwen3-VL-32B-Instruct, increasing by $+7.06$. These gains suggest that larger VLMs have the potential to exploit global spatial structure and self-tracking information, sometimes moving closer to the target even when final success remains limited. 
Overall, larger VLMs exhibit an emerging ability for self-localization and active planning. 
Yet absolute success rates remain below 10\%, even in this best-case setting, confirming that converting allocentric spatial information into long-horizon embodied action remains the central bottleneck.

\textbf{Global visual routes remain difficult to execute in embodied interaction.}
The \textit{Visual Route Map} variant provides the route as visual annotations, reducing the need to infer a route from scratch.
For Qwen3.5-27B, \textit{Visual Route Map} improves SR from 3.92\% to 8.24\%, suggesting that route annotations can support embodied navigation for larger models.
However, this improvement is not consistent across models: for Qwen3-VL-32B-Instruct, SR increases by only $+1.96$.
These results suggest that the bottleneck is not route planning alone, but also the concrete execution of a global route during embodied interaction.


\end{document}